# Ignorance-Aware Approaches and Algorithms for Prototype Selection in Machine Learning

Vagan Terziyan, Anton Nikulin

*Faculty of Information Technology, University of Jyväskylä, Finland,*
*e-mail: vagan.terziyan@jyu.fi; anton.nikulin@teamdev.com*

**Abstract:** Operating with ignorance is an important concern of the Machine Learning research, especially when the objective is to discover knowledge from the imperfect data. Data mining (driven by appropriate knowledge discovery tools) is about processing available (observed, known and understood) samples of data aiming to build a model (e.g., a classifier) to handle data samples, which are not yet observed, known or understood. These tools traditionally take samples of the available data (known facts) as an input for learning. We want to challenge the indispensability of this approach and we suggest considering the things the other way around. What if the task would be as follows: how to learn a model based on our ignorance, i.e. by processing the shape of "voids" within the available data space? Can we improve traditional classification by modeling also the ignorance? In this paper, we provide some algorithms for the discovery and visualizing of the ignorance zones in two-dimensional data spaces and design two ignorance-aware smart prototype selection techniques (incremental and adversarial) to improve the performance of the nearest neighbor classifiers. We present experiments with artificial and real datasets to test the concept of the usefulness of ignorance discovery in machine learning.

Keywords: data mining; classification; ignorance; prototype selection; adversarial learning

## 1. Introduction

*Empty spaces – what are we living for*

− Queen (1991). "*The Show Must Go On*"

Fighting data and knowledge imperfectness, from a slight uncertainty to a complete ignorance, is an important agenda for, e.g., Geographic Information Systems (GIS) for many years. Research on handling related issues has focused mainly on problems with spatial data and representations of spatial objects. Couclelis (2003), however, broaden the discussion on the ignorance in the geospatial domain by shifting the focus from information to knowledge and figuring out that a surprising number of things that we cannot know (or questions we cannot answer) are not the result of imperfect information. The author argues for accepting uncertainty and ignorance as natural and deep-rooted properties of complex knowledge, which need to be studied rather than excised. Leyk, Boesch & Weibel (2005) are sure that there is always something left one cannot know, and that a spatial analysis must not only explore what can be known but also improve our awareness of what cannot. Ignorance may have some common properties with the information we already know. According to De Bruin, S. (2008), sometimes a good assumption on a particular distribution of spatial features may help to recover some information about the missing ones. O'Sullivan & Udwin (2014) presented a framework on how to address the ignorance statistically by discovering a hypothesized spatial processes and assessing observed patterns against it. Mason *et al.* (2016) show that smart visualization of spatial ignorance may essentially improve human reasoning and decision-making. Yuan *et al.* (2005) argue that geographic data has unique properties, which require special consideration. For example, size, shape and boundaries of geographic objects can affect the data mining and automated knowledge discovery about the geographic processes, meaning that geographical objects cannot necessarily be reduced to points without information loss. We believe that the ignorance in GIS context has also similar meaningful properties (size, shape and boundaries), which is an additional opportunity for knowledge discovery rather than a thread.

It is known that voids or ignorance zones (in the topography sense) is a natural phenomenon even in the largest spatial datasets, and various void-filling algorithms are addressing it (Reuter, Nelson & Jarvis, 2007). Kinkeldey (2014) noticed that the uncertainty in remotely sensed imagery, resulted from such voids, plays an important and even positive role within the land cover change analysis and geovisual analytics. Combining the map view and the change info view allows visualizing and getting benefits not only from the discovered uncertainty spaces but also from their observed dynamics (Kinkeldey, 2014). In a larger scale, such as the astronomy or astrophysics, the ignorance is often associated with the *dark matter*, which is a hypothetical kind of matter that cannot be observed yet, but some (e.g., gravitational) effects of it to the visible matter can be captured to some extent and they help to infer not only the existence but also some unknown properties of the dark matter. Another interpretation of the ignorance in the same scale would be the *cosmic voids* or (almost) empty spaces between the largest structures in the Universe, which require special algorithms for identification and clustering (Chan, Hamaus & Desjacques, 2014). It is interesting that such ignorance-as-an-uncertainty (dark matter) covers the surface of the ignorance-as-a-

void (cosmic void) and that the properties of dark matter are changing according to radial distances from the void centers (Brunino *et al*., 2007).

DeNicola (2017) examines many forms of ignorance (ignorance as place, boundary, limit, and horizon) and argues that ignorance is more than just a void because it has dynamic and complex interactions with knowledge.

Ogata *et al.* (2013) consider knowledge transitions within four states: "(1) I know what I know", "(2) I know what I don't know", "(3) I don't know what I know" and "(4) I don't know what I don't know". They show that the transition from the "(4) I don't know what I don't know" state to the state "(2) I know what I don't know", supported by the technology, has a special importance and has been appreciated by the mobile learners. Therefore, knowing the boundaries of own ignorance is at least better than knowing nothing.

As one can see, the approaches related to handling ignorance are largely biased by a specific domain. We want, however, to study ignorance in more abstract terms to release it from the domain context. While addressing the ignorance-related abstractions, we are going to use the terminology of the data mining, machine learning and knowledge discovery in datasets.

In this paper, we provide some algorithms for the discovery and visualizing of the ignorance zones in two-dimensional data spaces and design several ignorance-aware smart prototype selection techniques to improve performance of the nearest neighbor classifiers. We experimentally prove the concept of the usefulness of ignorance discovery in machine learning. Therefore, the major addressed research questions are: how to approach the ignorance discovery in datasets; how to benefit from the discovered ignorance in supervised learning and classification.

The rest of the paper is organized as follows: in section 2, we discuss how the ignorance concept related to and can be useful for the machine learning with Open World Assumption; in section 3, we suggest several different approaches to define, capture and visualize the ignorance; in section 4 we present the generic model of ignorance discovery, which takes into account the distribution of data within the domain and the shape of the domain boundary; in section 5 we present one of possible use cases for the ignorance discovery, particularly two ignorance-aware algorithms for prototype selection (incremental and adversarial) in supervised instance-based learning, and we experimentally demonstrate the added value provided by the ignorance awareness (for our experiments we used eight datasets from the popular machine-learning repository); and we conclude in section 6.

## 2. Ignorance, AI and the Open World Assumption

Alan Turing (Turing, 1950) believed that a machine might pretend to be intelligent like a human if it provides appropriate and sustained responses to any questions in a similar to human manner. Much later, Warwick & Shah (2017) demonstrate that, actually, the truly intelligent machine is the one that knows also when and why be silent. Their experiments show that a machine sometimes passes the Turing test simply by not saying anything. Therefore, an absence of data in some subspaces of the data space possibly has some hidden meaning, and the knowledge discovery tool may attempt to find an answer on what the data collection system is silent about and why.

The good times are coming back when the Artificial Intelligence (AI) is making the major news in the technology world and not only pretending on perfectness in traditionally human intelligent games like Go and Chess (Silver et al., 2017), but also intimidating to push some human jobs away from the intelligent jobs' market (Ford, 2015). According to famous assumption made by Turing (1950), it is not feasible to hard-code a fully skilled AI. The only way is to program a "child" (basic intelligent instincts and capabilities to learn own skills) and then train it to the level it will be capable for further self-development. Such training (called Machine Learning) enables transformation of some observed or communicated evidence (data), either in its raw form or pre-processed/labeled by "teachers", into various forms of executable knowledge aka intelligent capabilities of AI systems. Several decades of evolution of the machine learning techniques brought us to the deep learning stage, when hidden patterns and useful features can be discovered from the raw heterogeneous data at various levels of abstraction (LeCun, Bengio & Hinton, 2015).

Data mining with related and popular AI tools for knowledge discovery, like machine (supervised and deep) learning, are about processing available (observed, known and understood) samples of data (also named as data points or data exemplars) aiming to build a model (e.g., a classifier) to handle data samples, which are not yet observed, known or understood. These tools can be very different in their way to learn and represent a model or a pattern discovered from the data, however, there is one thing, which makes all of them similar. They all take the available data (known facts) as an input for learning. We want to challenge the "evidence" of this statement and we suggest considering the things the other way around. What if the task would be as follows: "how to learn a model (e.g., a classifier) based on our ignorance, i.e. by processing the voids within the available data space?" Can we improve traditional classification by modeling also the ignorance? There is an excellent quota by Thomas Pynchon from his fiction book (Pynchon, 1984): "Everybody gets told to write about what they know … The trouble with many of us is that at the earlier stages of life we think we know everything …, we are often unaware of the scope and structure of our ignorance. Ignorance is not just a blank space on a person's mental map. It has contours and coherence, and … rules of operation as well. So as a corollary to writing about what we know, maybe we should add getting familiar with our ignorance…" Looks like quite a reasonable advice. We believe that, if the intelligence indicates the extent we recognize, understand, organize and use observed evidence, then the wisdom would be the extent we recognize, understand, organize and use our ignorance. Certainly, the ignorance does not have any data as known evidence for processing, but it has some shape at least, which is the boundary between the known (data samples) and the unknown (voids within the data). This gives us some hope due to the assumption: the "geometry" of the shape of the ignorance is a useful knowledge and it is a benefit to the traditional classifiers. In this paper, we want to check this assumption and provide some algorithms for discovering and visualizing the ignorance as well as to discuss and experimentally study some potential applications of the ignorance learning.

The prevailing majority of the machine learning algorithms process training sets as the "closed world" datasets. By following the Closed World Assumption, one concludes from a lack of some information about an entity in the dataset that this information is false (Reiter, 1981), which means that the unknown simply cannot be true by default. An alternative to it is the Open World Assumption, according to which a lack of information does not imply the missing information to be false. In the open world, negative data is listed explicitly in the dataset and queries may be either looked-up or derived from the data and the axioms (Minker, 1982). There is no assumption that certain data is considered false just because positive prove was not found. For many business problems, such as, e.g. finding a route based on the information about the available flights and cities, the closed world assumption works well because such domain implies the truth of negative facts (Reiter 1981). Inferring the result based on the absence of data might not be an optimal approach for more complex problems, such as classification of malignant tumors, where creating a comprehensive dataset is very expensive or simply impossible.

Consider the example in Figure 1. We have two groups of data samples in a 2D data space, the "green" ones and the "red" ones, which belong to two different classes. According to the closed world assumption (Figure 1 (a)), no other classes of data points (except the green ones and red ones) are possible. Therefore, a machine learning algorithm can draw the so-called decision boundary (linear one in this particular case of a classifier), which separates the whole space to two subspaces assuming that any point (a new one from, e.g. testing set) can be classified to either being green or red depending on which side (subspace) regarding the decision boundary it is located. However, if we accept the open world assumption (Figure 1 (b)), then we have to leave some space to the unknown yet data samples from some other classes than from the two ones we already know. Therefore, we have to divide the space, as shown in Figure 1 (c), so that we have a subspace for potentially more green points, a subspace for red points and a subspace (named "ignorance") where we may estimate points from one or more other (yet unknown) classes.

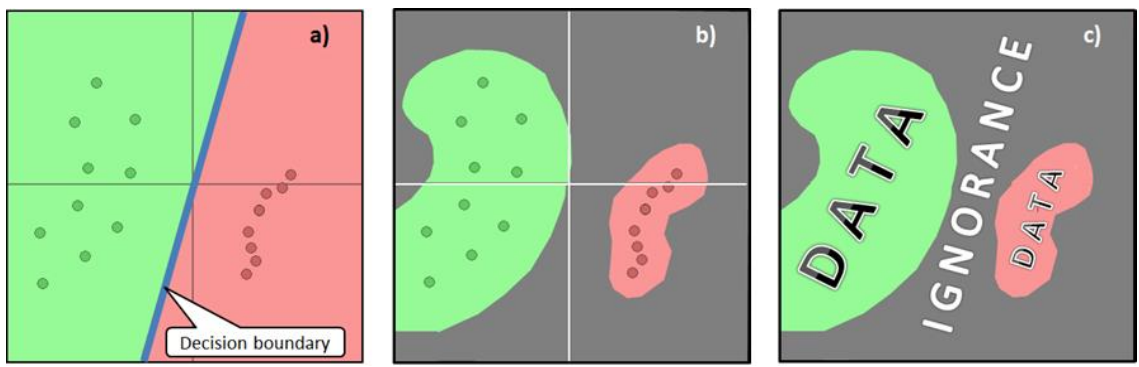

Figure 1: Supervised learning visualized for (a) the Closed World Assumption and (b-c) the Open World Assumption, which assumes the presence of the ignorance zones.

We believe that by applying not only the closed world assumption but also the open world approach, the existing machine learning algorithms could potentially make smarter decisions and improve their results. Instead of considering missing data as a false signal, the algorithms can benefit from the knowledge of unknown regions. The driving force of learning is the process of analyzing already available data and looking for areas where the dataset has the least amount of information. Such areas without data we name as *ignorance zones* (either data voids or confusion areas). These spaces do not give much benefit if taken separately, but their shape and size can help to understand accessible data better, make the analysis of it more efficient and even discover new classes of it. Therefore, we consider the *ignorance* to be a more generic concept comparably to the *uncertainty*. If (according to the Closed World Assumption) we fix the number of different classes in advance (like the two classes in Figure 1 (a)), then the classifier categorizes every point in the space with some level of uncertainty, which shows how confident is the classifier with its output. The closer the point to the decision boundary between the two classes the higher the uncertainty, up to the 50-50 at the decision boundary. In the case of the Open World Assumption, we can be as well uncertain at some points, which of the known classes must be applied there and with which level of confidence, however, in the same time we can be ignorant if any of the known classes is the right output or, probably, this particular point is the instance of the new, yet unknown class. Therefore, ignorance includes uncertainty as possible scenario together with potential confusion on the completeness of the available class labels set.

## 3. Ignorance discovery and visualization

### *3.1. Ignorance driven by Gabriel neighbors*

A considerable part of machine learning tasks (such as a supervised learning or classification) works with labeled data, in which every instance is attributed already or expected to be attributed to some class. Instances of the same class form a cluster that has certain unique characteristics that are different from other classes. As clusters possess different properties in Euclidean space they are commonly separated from each other by some kind of a void (Figure 1 (c)). When there are no data instances inside the void, there is no way to be confident on what might be hidden there. Such areas of emptiness located between known clusters represent the concept of ignorance or confusion zones.

In a two-dimensional Euclidean space, ignorance can be constructed from component ignorance zones, which are the largest empty circles touching some of the known points. A circle is chosen because the main property of it is that all the points of its boundary are equally distant from the center. The center of ignorance zone is called focus and it represents a place of the maximal confusion, especially if the circle touches several points from different classes. We consider two types of ignorance zones in 2D: based on two points and based on three points as shown in Figure 2. In the first case, the ignorance zone is an empty circle touching two heterogeneous (i.e., belonging to different classes) data points in such a way that the line segment between these two points is a diameter of the circle (Figure 2 (a)). The data points, which are the "parents" of such ignorance zones, are known to be the Gabriel Neighbors (Gabriel & Sokal, 1969). In the second case, the ignorance zone is an empty circle touching three heterogeneous and not collinear points (Figure 2 (b)). Although in some situations a circle can be built around four and more points (for example, around a square), it is not a generic case. Therefore, in 2D spaces we consider only pairs and non-collinear triples of data because it is always possible to make a circle around them.

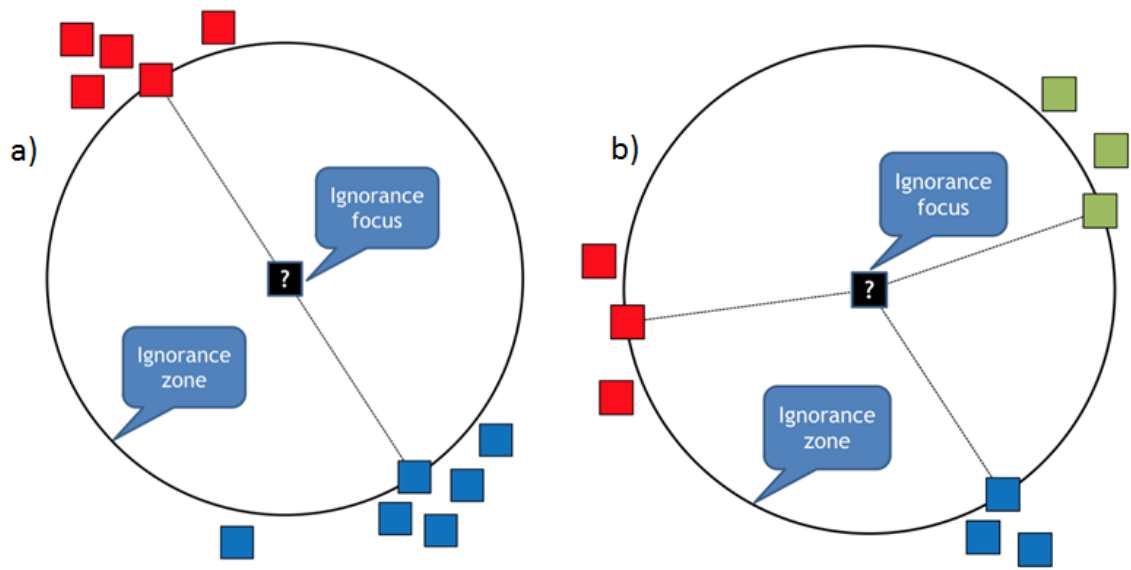

Figure 2. Two- (a) and three- (b) point ignorance zones and ignorance focuses.

In more generic $n$-dimensional spaces, we will have $n$ variations of the ignorance zones constructed by $2, 3, \ldots, n+1$ parent points and the zones will be the $n$-dimensional hyperspheres.

Discovered ignorance foci are valuable because a classification algorithm may be confused and tend to fail in these areas as they are close to decision boundaries where the transition from one class to another one happens.

Figure 3 presents few screenshots of our experiments with the artificial datasets where we used a simple algorithm to discover the two and three parent ignorance zones (red circles) in 2D space and their focuses (black points).

These zones and their foci presented in the figure are a kind of naïve model of curiosity, which is something like: "OK, I see your categorized facts, but can you answer few more questions on what is hidden here in the foci?" Realistic models of ignorance can be far more sophisticated as it will be shown in the following sections.

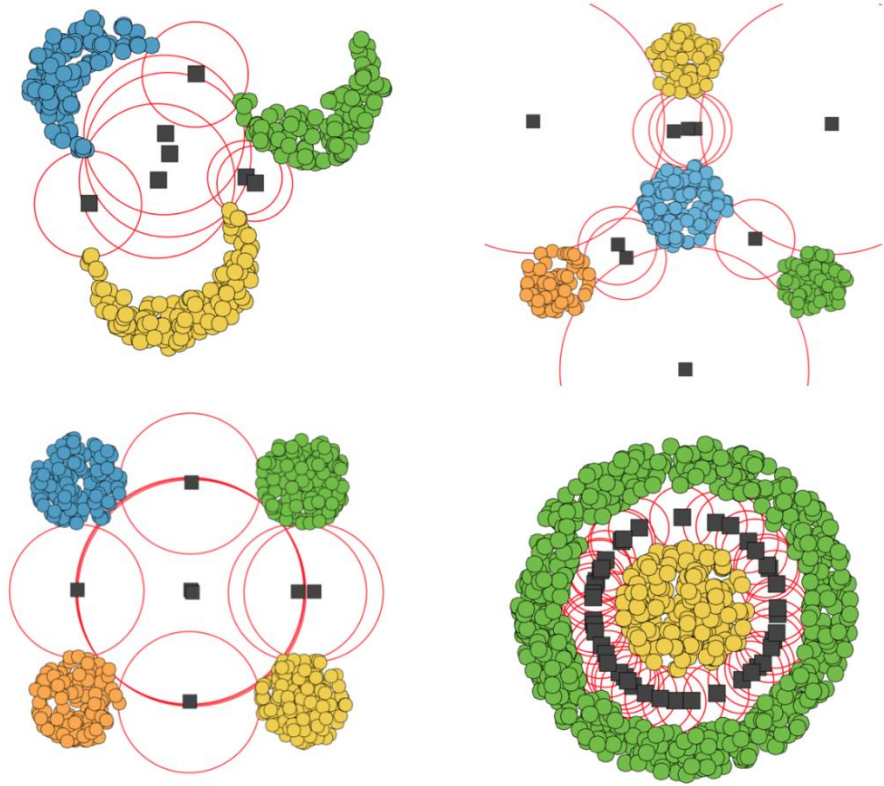

Figure 3. Screenshots of the discovered ignorance zones for the artificial datasets in 2D using Gabriel neighbors.

## 3.2. Shape-Aware Ignorance

Let us present an algorithm, which uses wider context (comparably to the previous algorithm) while discovering the ignorance zones. Assume that we have the set of labelled data, i.e. several differently colored manifolds in the data space depending on the amount of different labels (classes). We also assume that the whole shapes of the manifolds (not just the parts close to the decision boundary) may influence the contents and structure of the potential ignorance zones. To address this challenge we introduce an algorithm named "*k* Nearest and Different Neighbors" (*k*NDN).

*k*NDN is based on a following set of rules (in terms of a genetic algorithm):
- The term "cluster" is used to name each of the manifolds in the data space around the groups of the data exemplars labelled with the same class (color);
- All the clusters are potential parents and the discovered ignorance points are their children;
- All parents have as many "chromosomes" as they have data exemplars;

- Each pair of chromosomes from different parents may produce a child (exemplar of ignorance) and the child will be located exactly in the middle of the parent chromosomes;
- Each chromosome can be used only $k$ times for making children (then it will be "retired") and no more than once with the chromosomes from the same partner;
- The closer parent chromosomes are located to each other the faster they produce a child (if the distance between two pairs of chromosomes is the same then the advantage is given to more "fresh" (relaxed from previous birth) parents);
- Additional (optional) rule: If a newly born "child ("Ignorance Point") appears in the area of certain cluster of data (within one of parent's or some other one), then transform ("recolor") the ignorance point into the data point ("chromosome") of that cluster, which will have the full right to make own children as the regular data exemplar.

The algorithm gives definite advantage to nearest Gabriel neighbors to give birth to the ignorance points (like in the previous basic algorithm), however, due to special policies, it also gives a chance to more distant data exemplars from different manifolds contribute to the ignorance zones content creation. This makes a major difference with the basic algorithm from the previous sub-section, because now the whole shapes of the manifolds matter for the shapes of the ignorance zones between them.

The choice of the parameter $k$ influences the outcome of the $k$NDN algorithm in a similar way as this parameter influences the traditional $k$-NN classification algorithm, however, with some specifics. Assume that $C$ is the number of different classes within the data space and $N$ is the number of different data samples. The choice of $k$ in $k$NDN is limited between 1 and $C$-1, while in $k$-NN it is limited between 1 and $N$-1. The choice of $k$ is a trade-off between the computational resources (expenses grow when $k$ grows in both $k$-NN and $k$NDN algorithms) and the smoothness (noise resilience) of the discovered boundary, which is expected to be better with the bigger $k$. In $k$-NN, we are talking about the boundary between the classes of the data and, in $k$NDN, we are talking about the boundary between the data and the ignorance areas.

Figure 4 presents few screenshots of our experiments with the artificial datasets where we used $k$NDN algorithm to discover ignorance zones in 2D space.

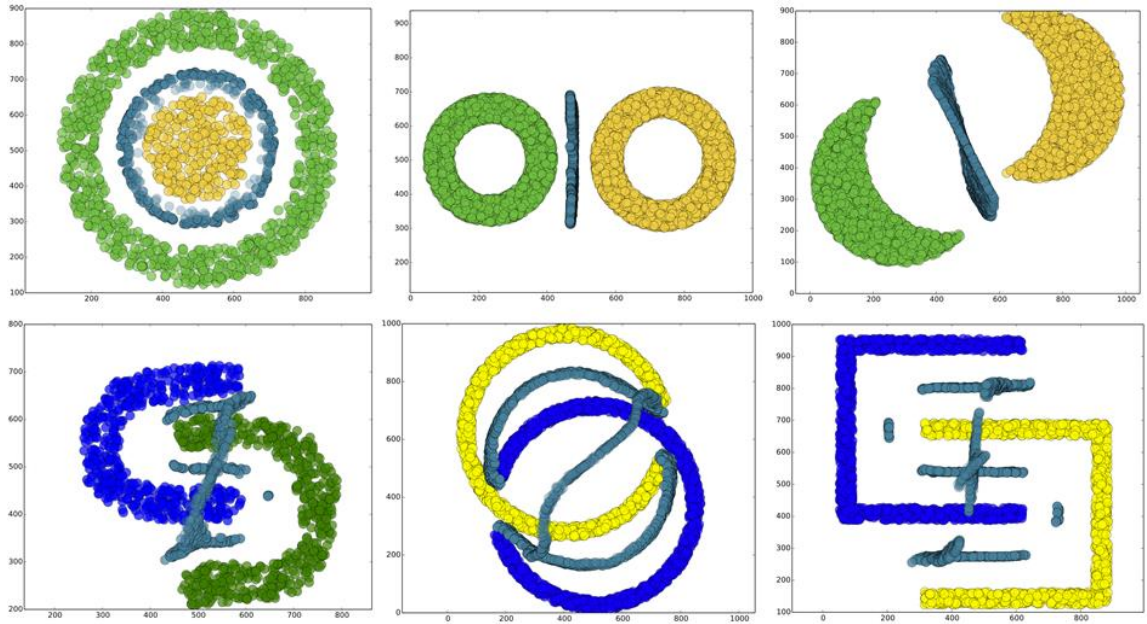

Figure 4. Screenshots of the discovered ignorance zones (set of grey points) for the artificial datasets (two differently colored data manifolds) in 2D with the *k*NDN algorithm.

As we already mentioned, the discovered ignorance zones can be treated as potential curiosity areas. Simple example in Figure 5 presents a situation from a popular "Battleship" game. In this game, players are more interested not in the ships ("cell clusters") discovered and "killed" already, but rather on empty spaces, where uncovered yet ships of the opponent might be located. Here, in the figure, one can see two already killed ships and now the decision is being made on where to shoot next, i.e. which parts of the remaining voids we are curious to explore with the next shot. If to use the *k*NDN algorithm and consider the two known areas around the killed ships as two different manifolds of data, then their valid children (potential curiosity points), "born" in a numbered order, can be seen in Figure 5. One possibly reasonable place to shoot next would be within this curiosity zone.

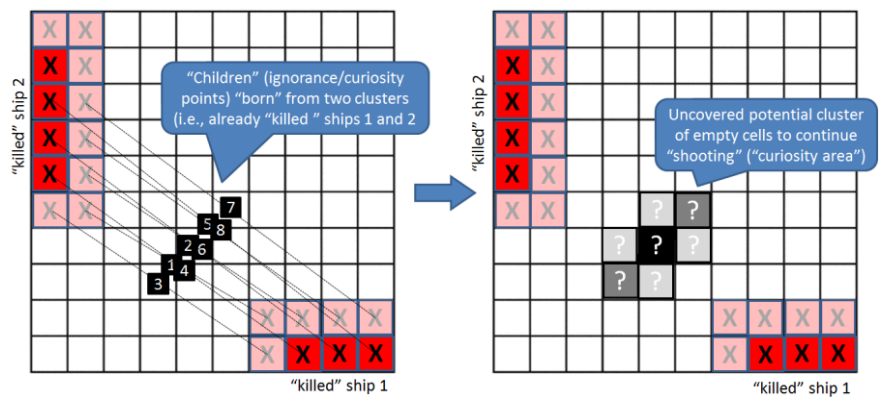

Figure 5. A simple illustration of the *k*NDN algorithm using the "Battleship" example.

Another possible way to discover ignorance zones with respect to the shape of the data clusters/manifolds boundaries would be use of the following distance measure. Assume we have a continuous boundary $\mathbb{A}$ around certain cluster of homogeneous (by color/lable) data exemplars a continuous boundary $\mathbb{B}$ around another cluster of data exemplars. Then the *Manifold Distance* between $\mathbb{A}$ and $\mathbb{B}$ would be as follows:

$$\forall A_i (A_i \in \mathbb{A}) \exists B_j (B_j \in \mathbb{B})[d\,(A_i, B_j) \leq k] \,\&$$
$$\&\, \forall B_r (B_r \in \mathbb{B}) \exists A_s (A_s \in \mathbb{A})[d\,(A_s, B_r) \leq k] \rightarrow$$
$$\rightarrow \mathfrak{D}\,(\mathbb{A}, \mathbb{B}) = \min(k), \text{ which means:}$$

$$\mathfrak{D}\,(\mathbb{A}, \mathbb{B}) = \max\left[\max\nolimits_{\forall A_i(A_i \in \mathbb{A})} \min\nolimits_{\forall B_j(B_j \in \mathbb{B})} d(A_i, B_j), \max\nolimits_{\forall B_j(B_j \in \mathbb{B})} \min\nolimits_{\forall A_i(A_i \in \mathbb{A})} d(A_i, B_j)\right].$$

Such distance measure has interesting physical interpretation in 2D space. Assume we have a "rope", one end of which is placed somewhere on the boundary $\mathbb{A}$ and the other end is placed somewhere on $\mathbb{B}$. Let the ends of the rope freely slide along the appropriate boundary if being pulled. The rope is capable of bending if necessary, but it is unable to stretch more than its original size without being broken. Assume we need to make a full round trip with the first end of the rope along the boundary $\mathbb{A}$ so that we are not breaking the rope. Then, we want a similar (safe for the rope) trip with the second end of the rope along the boundary $\mathbb{B}$. The shortest lengths of the rope needed to make all these possible would be the Manifold Distance $\mathfrak{D}\,(\mathbb{A}, \mathbb{B})$. Such measure essentially depends on the shapes of the manifold boundaries; therefore, it brings an interesting flavor also to the appropriate ignorance zones discovery.

Ignorance discovery between two manifolds is illustrated in Figure 6 and it goes according to the following steps:

1) first, the Manifold Distance $\mathfrak{D}\,(\mathbb{A}, \mathbb{B})$ is computed (Figure 6 (a));
2) the valid pairs of "parents" $(A_i, B_j)$ from the manifolds' boundaries are nominated as follows: $\forall A_i, B_j \{A_i \in \mathbb{A};\ B_j \in \mathbb{B}; d(A_i, B_j) = \mathfrak{D}\,(\mathbb{A}, \mathbb{B})\}$, where $d$ is distance between points in, e.g. Euclidean distance (see Figure 6 (b));
3) for each pair, the "child" (ignorance boundary point) is created, which is located exactly in the middle between parent points (Figure 6 (b-e));
4) the points of discovered ignorance boundary are connected to form an ignorance zone as shown in Figure 6 (f).

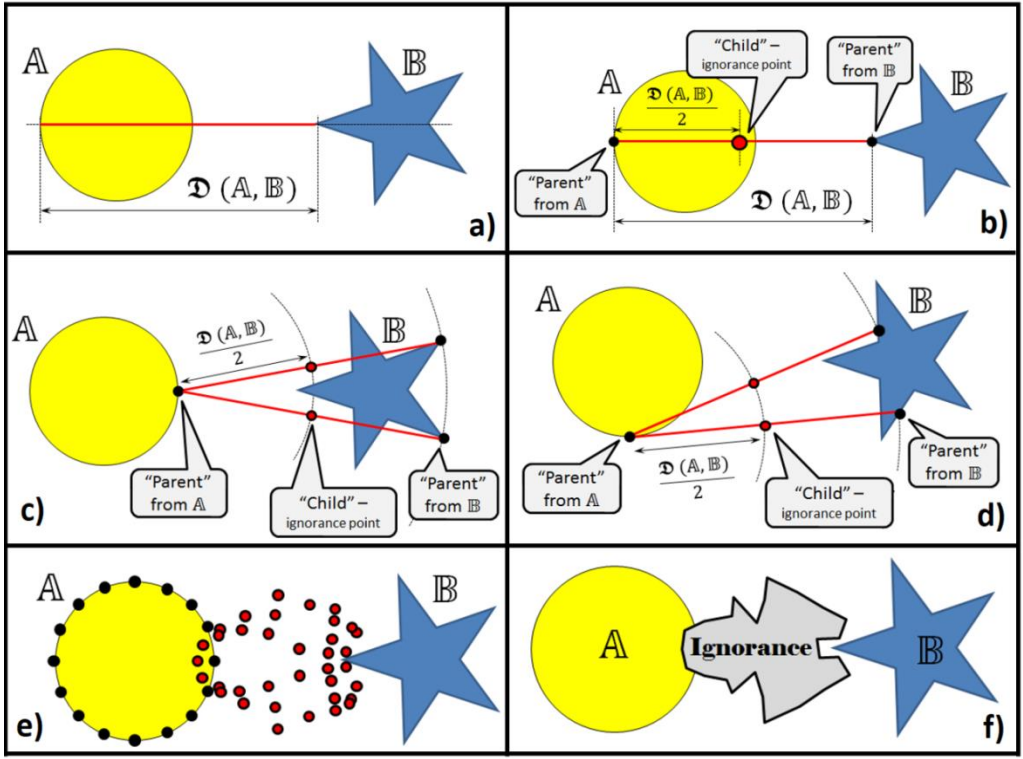

Figure 6. Illustration of the ignorance zone discovery with the Manifold Distance. One can see that the shape of such zone does not depend much on near-neighbors from both manifolds but more on overall shapes of the manifolds.

Consider yet another way to compute the ignorance points using the whole shapes of the manifolds in 2D. We name it as a *Balanced View* method. It allows computing the ignorance curves (aka decision boundaries) between the manifolds. The method is illustrated in Figure 7. For a couple of manifolds $\mathbb{A}$ and $\mathbb{B}$, the following rules are applied (in terms of a genetic algorithm):

1) We nominate the valid pairs of "parents" $(A_i, B_j)$ from the manifolds' boundaries $\forall A_i, B_j (A_i \in \mathbb{A}; B_j \in \mathbb{B})$ such that (for each pair) there exists an empty circle touching the manifold boundaries $\mathbb{A}$ and $\mathbb{B}$ exactly in the points $A_i$ and $B_j$ respectively (see Figure 7 (a));

2) we set up the "sightline" through each pair of the points $A_i$ and $B_j$ and we discover the corresponding points $A'_i$ and $B'_j$ $(A'_i \in \mathbb{A}; B'_j \in \mathbb{B})$ on the manifolds' boundaries so that sightline segments $A_i, A'_i$ and $B_j, B'_j$ are placed completely within the corresponding manifolds as shown in Figure 7 (a);

3) for each pair of parents $A_i$ and $B_j$ and with respect to their "counterparts" $A'_i$ and $B'_j$, the "child" (ignorance curve point) $I_k$ is created, which is located on the

sightline between the parents so that the following balance is kept (Figure 7 (a)):

$$\frac{d(A_i, I_k)}{d(B_j, I_k)} = \frac{d(B_j, B'_j)}{d(A_i, A'_i)};$$

4) the ends of the ignorance curve are computed as shown in Figure 7 (b). The sightline $A_i, B_j$ as well as the sightline $A_r, B_s$ a correspond to the circles with the infinite radius touching both manifold boundaries and therefore the children $I_k$ and $I_t$ are produced just in the middle between the parents without the use of counterparts;

5) all the children after being discovered (Figure 7 (c)) finally form the ignorance curve between the manifolds (Figure 7 (d)), which is a kind of shape-aware decision boundary between the manifolds.

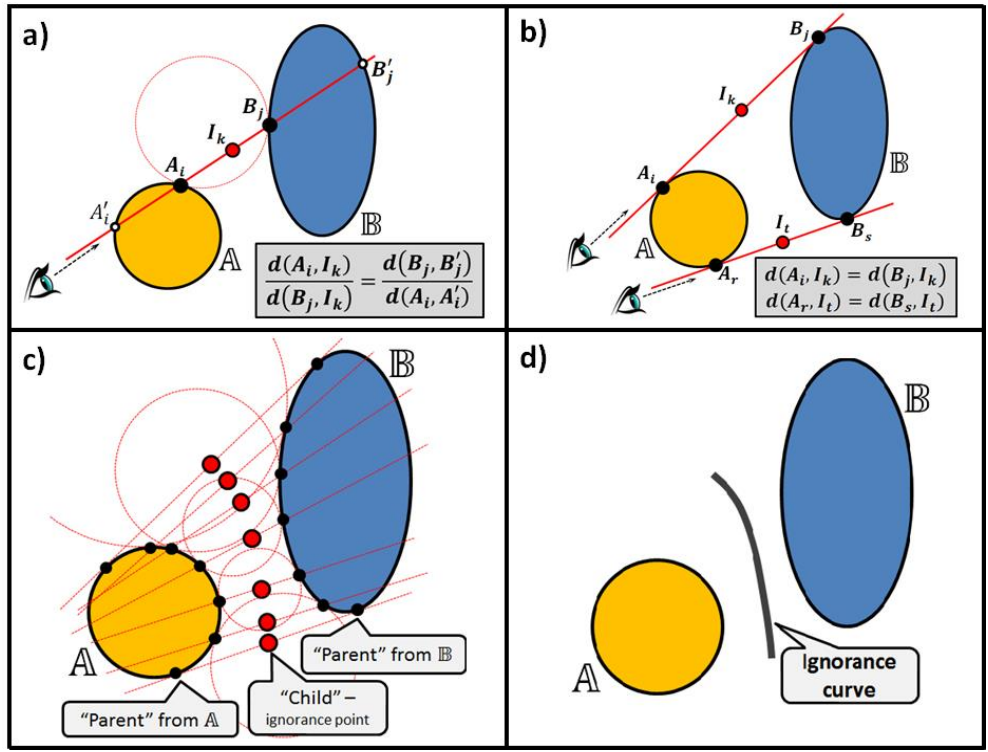

Figure 7. Illustration of the ignorance (decision boundary) curve discovery with the *Balanced View* method. Here the actual shapes of the manifolds influence the curve shape.

### *3.3. Density-Aware Ignorance*

Most of the previous cases were based on the Euclidian metric used to find the place of an ignorance point just in the middle between the two differently labelled (parent) points. Let us apply an alternative metric for the ignorance zones discovery. One of possible alternative metric would be the *Social Distance Metric* (Terziyan, 2017), which can be used on top of any traditional metric. For a pair of samples *x* and *y*, it measures a kind of "social asymmetry" among them by averaging the two numbers: the place (rank), which sample *y* holds in the list of ordered nearest neighbors of *x*; and vice versa, the rank of *x* in the list of the nearest neighbors of *y*. This specifics allows taking the density of points around *x* and *y* into account when measuring the distance between them. Figure 8 illustrates how this metric may effect on the ignorance point discovery.

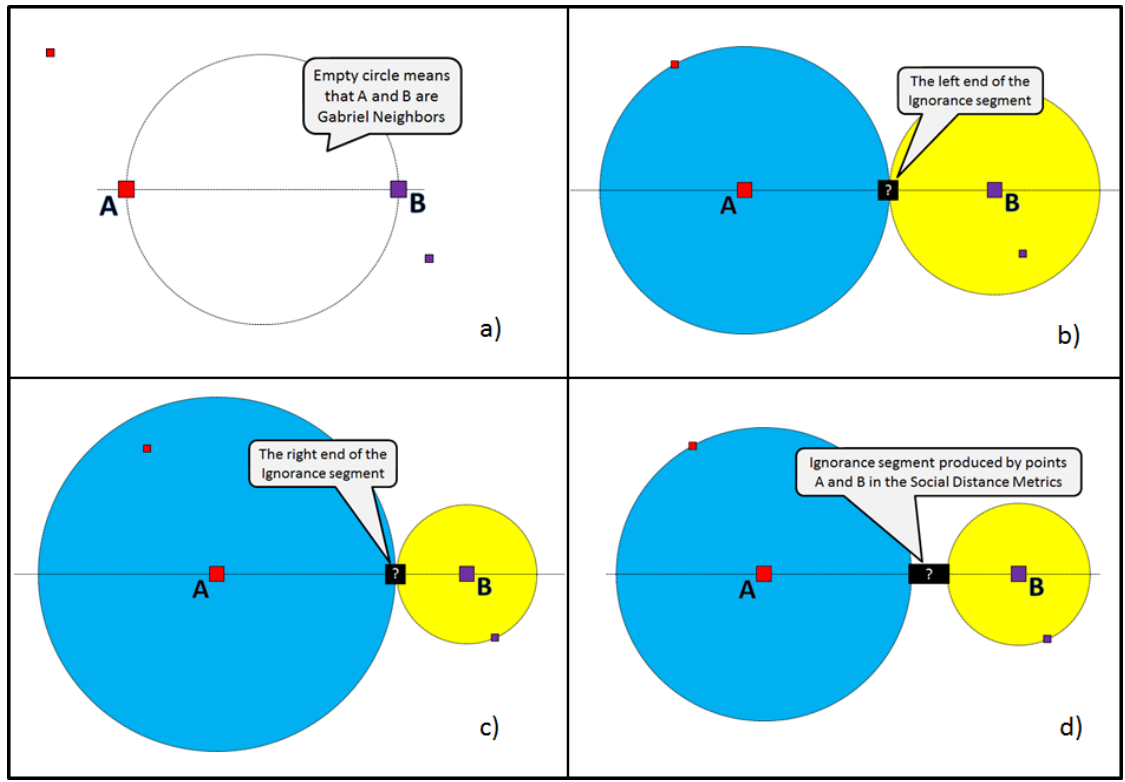

Figure 8. Illustration of the fact that the ignorance focus produced by two points A and B in the Social Distance Metric is not just a unique middle point between A and B but it is a line segment. One can see that the requirement for the ignorance line segment between A and B is that that amount of homogeneous data points within the blue circle around A is the same as the amount of data points within the yellow circle around B.

First (Figure 8 (a)), we find the suitable parents A and B for the ignorance point (i.e. the Gabriel neighbors). According to the *Social Distance Metric*, the intended "middle" point "?" would be the one with the following properties: it is located on the line connecting A and B; the circle centered in point A and touching the point "?" and circle centered in point B and touching the point "?" have the same amount of homogeneous (by color/label) data points (exemplars) within them. In a generic case, either one or several (infinite number within some interval) points may fit such a rule. Therefore, the result for A and B will be a line segment. Figure 8 (b) shows the existence of the left end of such a segment, and Figure 8 (c) shows the existence of the right end of the segment. The resulting ignorance zone (line segment) for A and B (as parents) is shown in Figure 8 (d).

The earlier considered algorithms (e.g., the basic Gabriel neighbor-based one or the *k*NDN) may use the *Social Distance Metric* and, instead of ignorance point as a "child" of two parents, may produce the ignorance line segments bringing new social density-driven flavor to the ignorance zones discovery.

## 4. A generic model of ignorance

Previous sections give a brief overview of some possible approaches to discover and shape the ignorance in datasets. In this section, we set up some basic policies related to ignorance definition and discovery, aiming to make more generic and practically useful approach, which will take into account not only the shapes of the data manifolds but also the shape of the domain boundary.

Assume we have some set of data labelled with some set of class tags (e.g., color). Let us consider the concept of a *domain* as a space with the same dimensionality as the data, such that: (a) all the data points (exemplars) from our dataset are located within this space and they will be used for training, validation and testing the hidden classification model; (b) the potential curiosity points (future queries to the discovered model during the use of it) are also expected to be within the space. Assume also that the data is preprocessed (normalized) so that all the data points can be placed compactly within a hypersphere (equidimensional with the data) with the boundary surface $D$. Let the hypersphere be centered in $O(D)$ with the radius $R(D)$ or simply $R$. One can use simple and fast algorithm to discover the bounding hypersphere for a set of points, see, e.g., (Elzinga & Hearn, 1972) or Ritter's bounding sphere algorithm (Ritter, 1990), which guarantees close-to-optimal solution in a reasonable time. One among other good reasons to choose a spherical shape for the domain is that the surface function of a hypersphere is differentiable at every point, which is not the case for a hyperrectangle domain in their corner points.

We argue that the ignorance (conflict, confusion, curiosity, etc.) zones within the domains of the datasets exist not only between the differently tagged data manifolds (as well as between heterogeneous individual data points) but also between these manifolds (as well as separate points) and the domain boundary, which can be considered itself as a kind of decision boundary between known and unknown.

To make it easy to present and visualize such a duality of the ignorance, we come back to the 2D data spaces. We consider two types of voids within any 2D domain (i.e., a circle), which can produce the ignorance zones: (a) empty circles centered on the domain boundary and touching at least one data point; (b) empty circles placed completely within the domain and touching two differently tagged data points.

Every void produces an ignorance zone, which is a circle centered at the same point as the void itself and having the radius computed following the basic rule:

$$\frac{Radius\ of\ the\ ignorance\ zone}{Radius\ of\ the\ void} = \frac{Radius\ of\ the\ void}{Radius\ of\ the\ largest\ possible\ void\ in\ the\ domain}.$$

The largest possible void may be an empty circle centered at some point on the domain boundary and touching the domain point located exactly on the opposite side of the domain boundary. Therefore, the radius of such void would be equal to $2 \cdot R$, where $R$ is the radius of the domain boundary surface $D$. If to denote the ignorance zone radius as $\varepsilon$ and the void radius as $r$, then, following the proportion above, we get the following basic formula for an ignorance zone size:

$$\varepsilon = \frac{r^2}{2 \cdot R}.$$

Figure 9 illustrates the origin of all the ignorance zones around the unique data point $A$, which is located somewhere within the domain $D$ so that the distance between $O(D)$ and $A$ is equal to $\varphi$. All the ignorance zones (Figure 9 (a-b)), when merged (see Figure 9 (c)), form a solid *area of ignorance* around point $A$. With a simple geometry (Figure 9 (d)), one may see that the ignorance zones remain some space untouched (we name it as a *believed certainty area*), which is an ellipse centered in $A$ and having parameter $a$ as semi-major axis and $b$ as semi-minor axis computed as follows:

$$a = \frac{h}{2 \cdot R} \cdot (2 \cdot R - h); \text{ and } b = \frac{h^2}{2 \cdot R}, \text{ where } h = a + b = \sqrt{R^2 - \varphi^2}.$$

Therefore, the closer the point $A$ to the domain boundary, i.e., $\varphi \to R$, the smaller is the believed certainty area, i.e., $a \to b \to 0$. On the other hand, the closer the point $A$ to the domain center, i.e., $\varphi \to 0$, the closer the believed certainty area would be to the circle with the radius, which is half of the domain radius, i.e., $a \to b \to \frac{R}{2}$.

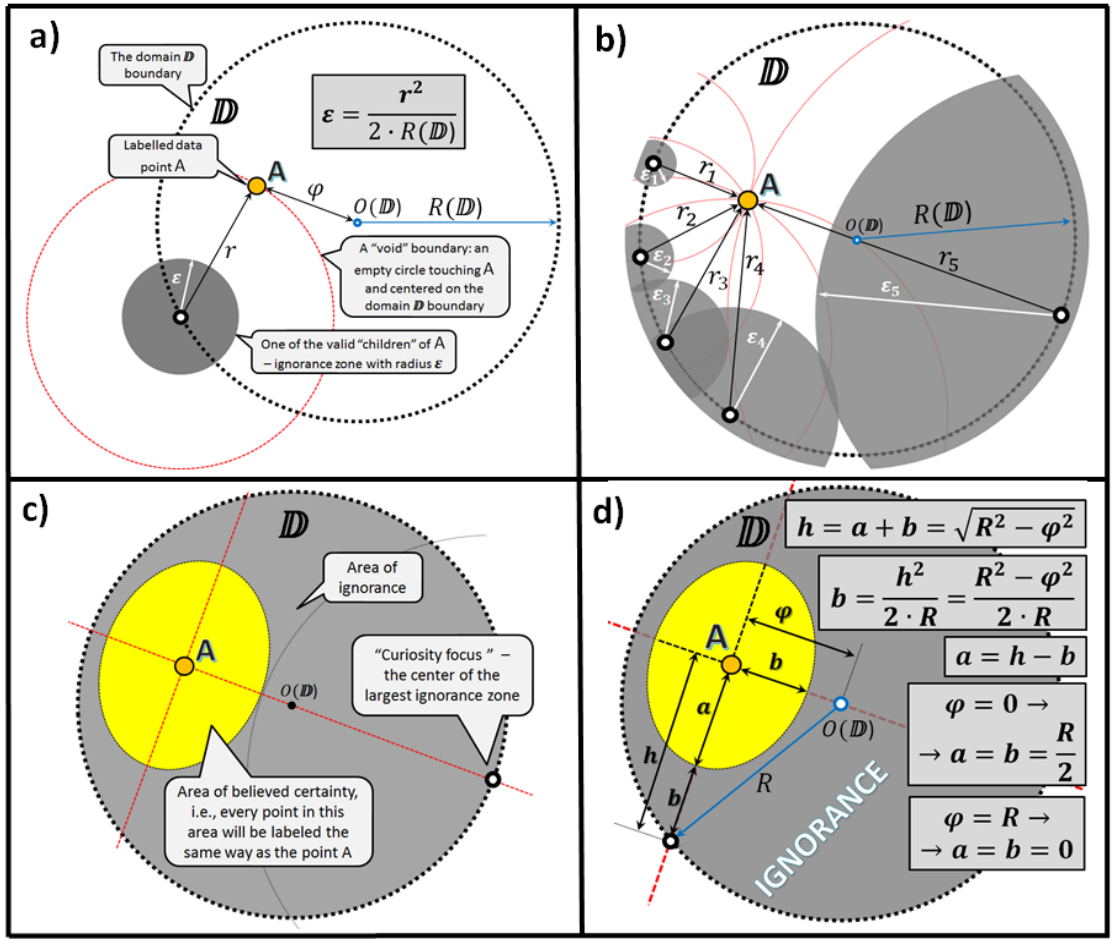

Figure 9. Illustration of the origin and parameters of the believed certainty and the ignorance areas around point A within the domain: (a) ignorance zone produced by one void; (b) ignorance zones produced by some other voids; (c) ignorance zones are merged so that the clear (elliptic) boundary is seen between the ignorance and certainty areas; (d) computations for the boundary between the ignorance and certainty.

Notice also the *curiosity focus* (Figure 9 (c)), which is the center of the largest ignorance zone. Assume we may ask from someone about the tag (class label) for a potential data point located exactly within the curiosity focus and get a correct answer, then the largest ignorance zone will collapse and we will get the maximal possible impact on the certainty-ignorance ratio within the domain.

Figure 10 illustrates the origin of all the ignorance zones created due to the conflict (different tags or class labels) between two data points *A* and *B* within the domain. Here in Figure 10, as well as in the previous case shown in Figure 9, we have to consider circular voids (empty circles) centered on the decision boundary, however, in the case of two conflicting points, the decision boundary is not only the domain boundary, but it is also the line, which goes via the points equidistant from *A* and *B*. All such voids, which touch both *A* and *B* and placed completely within the domain, make additional ignorance zones marked by red color in Figure 10 (a) and sized similarly as the previously seen voids created due to the conflict between a unique data point and the domain boundary (Figure 9). If to combine/merge both types of voids and their ignorance zones (Figure 10 (b)), then we will get the solid area of ignorance together with the two spots of believed certainty (Figure 10 (c)). In the yellow spot around the point *A*, we believe that all the points (potential test queries) will be tagged (classified) the same as the data point *A*; and, in the blue spot around point *B*, we believe that all the points will be classified to the same class as the data point *B*.

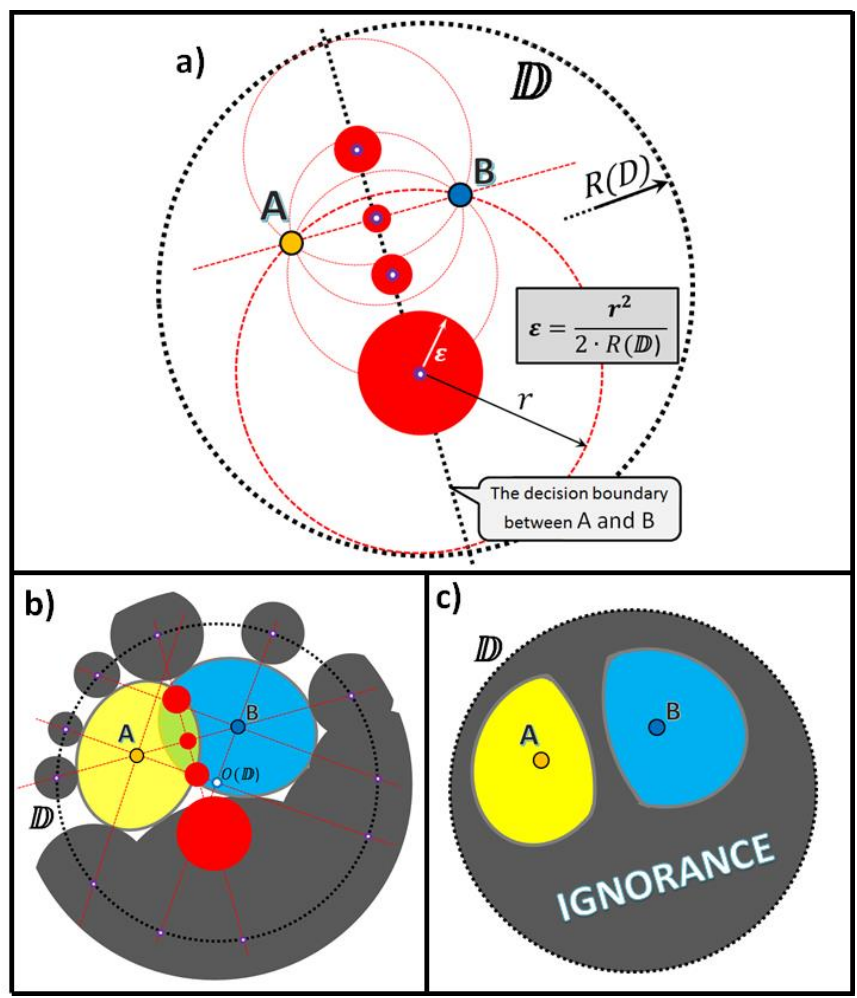

Figure 10. Illustration of the origin and parameters of the believed certainty and the ignorance areas created by a couple of conflicting data points *A* and *B* within the domain: (a) the conflict between *A* and *B* makes a decision boundary between them, where all the additional ignorance zones are centered; (b) the discovery of the *A-B*-conflict-related ignorance zones may go synchronously with the *A*-Domain-conflict-related and *B*-Domain-conflict-related ignorance zones discovery; (c) both types of ignorance zones are merged so that the clear boundary is seen between the ignorance and two heterogeneous certainty areas.

Therefore, we may see that, before discovering the potential ignorance zones within the data, we need to find all the decision boundaries around every data point, which separate each point from its neighbors. Such boundaries form so called "cells" or convex polygons around each point, which are usually visualized with a *Voronoi diagram* (Aurenhammer, 1991) in 2D spaces. Traditional Voronoi diagram represents a plane (usually Euclidean) with the data points as a set of cells such that each cell contains exactly one (generative) point, and every point in a given cell is closer to its generating point than to any other in the dataset. In our case, some cells, which are close to the circular domain boundary, may have some of the edges as arcs (not straight lines).

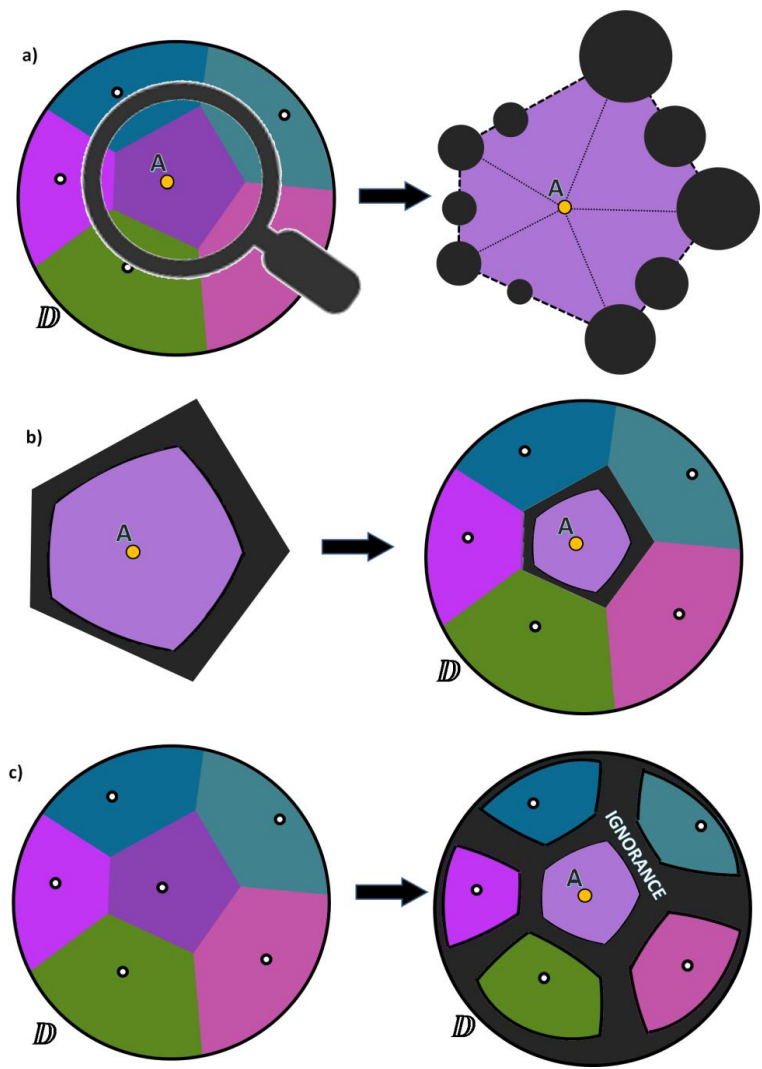

Figure 11. The example of the Voronoi diagram evolution towards the ignorance-aware Voronoi diagram: (a) the ignorance zones are computed separately for each Voronoi cell, which works as a small domain with generating point in it; (b) the ignorance zones within each cell are merged into the ignorance area around the point leaving also some space for a zone of believed certainty; (c) the same is done for all the Voronoi cell resulting to the ignorance-aware Voronoi diagram.

In Figure 11, we see an example of the domain with six differently tagged data points located within it and represented with a Voronoi diagram. Let us take one of the cells to start with (Figure 11 (a)). This cell looks just as a kind of domain containing one point. Therefore, we can assume that all potential ignorance zones that are based on the position of this point in its immediate vicinity will be centered on the Voronoi cell boundary, and the size of these zones can be calculated using the technique described above. The scaling factor $R(\boldsymbol{D})$, which is the size of the global domain $\boldsymbol{D}$, remains the same for all the computations. All such zones when merged (Figure 11 (b)) form an ignorance area around each point within each cell. Finally, we get a kind of "ignorance-aware" Voronoi diagram, in which the ignorance zones and the zones of believed certainty are clearly indicated (Figure 11 (c)).

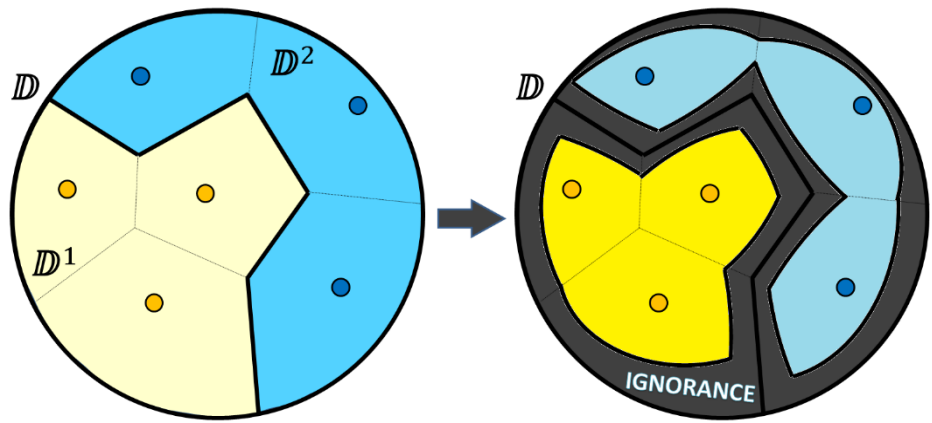

Figure 12. Ignorance-aware Voronoi diagram creation with two homogeneous clusters of data points: (on the left) the Voronoi cells with the same class label are merged into two subdomains separated by the decision boundary; (on the right) the ignorance zones are computed separately for each of two subdomains and merged into the ignorance area around the homogeneous groups of points leaving also some space for two zones of believed certainty.

In the real datasets, we may have groups of points with the same label (class tag). If some groups of such homogeneous points are located in the immediate vicinity of each other, then we have to merge their appropriate Voronoi cells into more complex subdomains than individual Voronoi cells. Consider similar example in Figure 12, which looks like the previous one from Figure 11, but this time our six points are not mutually heterogeneous and they form two homogeneous clusters (three yellow and three blue data points). One may see that not all the Voronoi boundaries (thin lines) are actually the decision boundary (bold lines), which either separates yellow from blue areas or it is the domain boundary. Therefore, we can merge three yellow cells into one complex cell (subdomain $D^1$) and also we can merge three blue cells into one complex cell (subdomain $D^2$). After that (see Figure 12), it is possible to use the same ignorance discovery technique separately for each subdomain. This will result to three bounded areas within the domain: yellow area of believed certainty such that every point within this area must be classified as a yellow one; blue area of believed certainty such that every point within this area must be classified as a blue one; and the ignorance area, in which we doubt on the class label.

Considered generic model of ignorance and its discovery technique can be generalized from 2D to n-D domains. In this case, instead of circles we will be dealing with hyperspheres and instead of the decision boundary as a line/curve we will be dealing with the hyperplane/surface. The resulting ignorance areas will be formed by merging the ignorance zones (hyperspheres, which are centered on the decision and domain boundaries and which radius is computed with same formula as for 2D consideration above). Naturally, all the ignorance discovery techniques will work similarly also with the hyperrectangle-looking domains, if needed.

## 5. Using ignorance as a driver for prototype selection

In this section, we try to address the potential question from the pragmatists on how one may benefit from the ignorance awareness. We take one of the classical problems in machine learning, which is prototype selection, and we experimentally study the benefits, which a smart use of the ignorance awareness provides for that.

In his recent study, Buchanan (2018) compares various approaches to forecasting in atmospheric physics in general and to weather forecast in particular and he argues that the "better predictions could be made by putting more focus on what we don't know, and possibly cannot know". This means that having huge collection of data and knowledge about a dynamic phenomenon still would not be enough for a reliable prediction, and there is a need to explore the ignorance zones of it. There more data we select from the collected measurements as a prototype for building the prediction model, the more risk of potential overfitting we may have, and, therefore, a smart prototype selection for the model construction must take into account also the boundaries of the ignorance zones to avoid focusing on possible noise within the data.

In supervised learning, classification is one of the most widespread techniques that help to categorize new previously unseen observations. After learning from a training dataset that contains a collection of labeled examples, a machine-learning algorithm tries to predict a class for a new unlabeled example during testing. A large number of classification algorithms use the distance between the new input examples and stored labeled examples when predicting the class label for the new ones. Such algorithms, like, e.g., the popular *k*-Nearest-Neighbor classifier (*k*-NN), which are called instance-based or lazy learners (Brighton and Mellish 2002), may not need a classification model built in advance, but they are capable of making decisions completely relying on existing prototypes from the training set. The *k*-NN logic is simple: for each new unlabeled example it finds *k* nearest labeled neighbors from the training set and chooses the most common class label according to the majority vote rule. Being an effective classifier for many applications, *k*-NN still suffers from multiple weaknesses (Kononenko & Kukar, 2007): high storage requirements as all the labeled examples are kept in the memory; demand for powerful resources to compute distances between a new input and all the original prototypes; low noise-tolerance because all the data instances stored in memory are considered relevant and possible outliers can harm classification accuracy.

The excellent review, the taxonomy and empirical comparison of the prototype selection algorithms is available in Garcia, Derrac, Cano & Herrera (2012). Taking into account many conflicting factors affecting the quality of the prototype selection, they noticed that the conclusion cannot be determined on the best performing method and the choice depends on the particular problem settings. The importance of noise filtering as one of prototype selection objective is currently discussed in Gupta & Gupta (2018). They argue that possible overlapping of the classes in datasets may be an indication of noise, and they proposed a new overlap measure aiming to detect the noisy areas in data. Among recent updates in the prototype selection techniques there is also article from Olvera-López, Carrasco-Ochoa & Martínez-Trinidad (2018), where they enable nominal features in the data and suggested considering two groups of selected instances: relevant and border prototypes. They argue on the effectiveness of their method also with the large datasets. A review focused more on prototype selection for 1-NN classifiers is recently provided in Zubek & Kuncheva (2018). They noticed that the compromise between cognitive psychology and machine learning in prototype selection would potentially be the best approach to the interpretable nearest neighbor categorization.

In this section, we want to make our small contribution to the prototype selection approaches by considering a so-called *curiosity-driven approach*, in which the discovered ignorance zones within already selected prototypes indicate iteratively the "curiosity focus" as a demand for selecting every new prototype from the dataset. The curiosity focus, which contains coordinates of the largest ignorance zone center, is used as the nearest neighbor query (see, e.g., Chen & Lu (2008)) to the original dataset.

We suggest two curiosity-driven and ignorance-aware algorithms for prototype selection, which use the ignorance zones discovery and we experimentally checked the performance of both of them.

*5.1. Incremental prototype selection*

In this paper, we present an incremental algorithm for prototype selection within 2D domains. We assume that the domain contains complete dataset inside it and represents a bounded space circumscribing all the available data instances. In our experiments, the rectangular and circular domains are validated and compared with each other. In the beginning, our algorithm estimates the boundaries of known data and computes the domain size and location. Therefore, we start each experiment only with the known domain boundary, and the space within the domain boundary is considered as an initial ignorance zone. Notice that the domain radius $R$, which is used for the ignorance size estimation, must be computed as: $R = \frac{\sqrt{X^2+Y^2}}{2}$ for the case of a rectangular domain where $X$ and $Y$ are the vertical and horizontal sizes of the rectangle.

For any form of the domain, the Incremental Prototype Selection (IPS) algorithm stays identical and works as follows:

1. The domain boundary for the given training set is calculated (either the smallest rectangular or the circle circumscribing all the points in the original dataset). The set of already selected prototypes is initialized as an empty one.
2. At every iteration step, the ignorance zones are discovered within the domain populated only with the already selected prototypes. (Notice that the domain as whole, being empty at the very beginning, is considered as the largest and the only ignorance zone at the initial stage of the iteration process).
3. At every iteration step, a curiosity focus (the center of the largest ignorance zone) is discovered; a nearest neighbor query is initiated to the original dataset; the data sample (closest to curiosity focus and located within the space of the corresponding ignorance zone) from the original dataset is taken and added to the already selected prototypes set. If the largest ignorance zone does not have new points to be taken from the original dataset, then the second largest ignorance zone is examined, and so on. Stopping criteria: the algorithm stops when: either none of the ignorance zones have vacant points in the original dataset; or the radius of the largest ignorance zone reaches some predefined minimum $\varepsilon_0$. Otherwise, the stages (2) and (3) are repeated continuously.

## *5.2. Adversarial prototype selection*

Adversarial learning in general and Generative Adversarial Networks in particular has recently become a popular type of deep learning algorithms and they have achieved great success in producing realistically looking images and in making predictions. Unlike the discriminative models, where the high-dimensional input data is usually mapped to a class label, the adversarial networks offer a different approach where the generative model is pitted against the discriminative one (Goodfellow *et al.*, 2014). The adversarial method works as a system of two neural networks, where one of them, called the generator, produces fake prototypes of the intended class and the other one, called the discriminator, tries to uncover the fake by evaluating how well the prototype fits the distribution of real prototypes of this class. During the process of training, both networks are co-evolving up to the perfectness in performing their conflicting objectives. In this paper, we are not going to copy the generative adversarial networks as such, but rather use the idea of two models (two selected prototype sets in our case), which are improving each other while competing against each other; and we apply the idea to the previously described concept of ignorance-aware prototype selection.

We use one of the possible models of interaction between a student and a professor related to the learning outcomes assessment process as an abstract analogy to describe our adversarial algorithm. Assume that some (strict and thorough) professor aims to assess impartially the knowledge of the (lazy) student, whose main goal is to answer questions correctly and pass an exam with the minimal study effort. This symbiosis results into the mutual benefit: the professor learns, which knowledge to assess and how to detect the gaps (i.e., student's ignorance) with the least number of questions, and the student learns how to study the least amount of information, but, in the same time, how to be capable to address the professor's questions as correctly as possible. Such interaction between a student and a professor resembles a classification problem where a model, which is based on carefully selected training set, aims to answer questions (classification queries), asked from a carefully assembled testing set. In this case, we get not only a well-trained classifier but also an adequate evaluation of it based on the most challenging tests. The idea of the Adversarial Prototype Selection (APS) algorithm, which we present in this paper, fits well the concept of a generative adversarial network (Goodfellow *et al.*, 2014), i.e., assuming that the student acts as a generative model and the professor acts as a discriminative one.

The APS algorithm supposes that the two competing actors fill their prototype sets incrementally and independently but with one major assumption: at each iteration, the professor is aware on the ignorance area within the student's prototype set (such awareness is used for selecting hard questions for the exam); and, vice-versa, the student will be always aware on the ignorance area of the professor's prototype set (such awareness is used for minimizing the effort to prepare for the exam).

For any form of the domain, the Adversarial Prototype Selection (APS) algorithm stays identical and works as follows:

1. The domain boundary for the given training set is calculated (either the smallest rectangular or the circle circumscribing all the points in the original dataset). The sets of already selected prototypes are initialized as the empty ones for both authors: the professor and the student.
2. At every iteration step, the ignorance zones are discovered (in a similar way as in the IPS algorithm) synchronously and independently for the professor using his/her already selected prototypes and for the student using his/her already selected prototypes.
3. At every iteration step, the curiosity zones are discovered for both actors separately; for the professor:

$$Curiosity_{professor} = Ignorance_{student} \cap Ignorance_{professor};$$

and for the student:

$$Curiosity_{student} = Ignorance_{student} \cap (\neg Ignorance_{professor});$$

(*Notice that initially the student's curiosity zone will be empty and the professor's curiosity zone will be the whole domain, which means that the professor will be the first one to start recruiting the prototypes from the dataset*);

4. At every iteration step, a curiosity focus (the center of the largest circle within the curiosity zone) is discovered for both actors separately; appropriate nearest neighbor queries are initiated to the original dataset (from the professor and from the student); the data sample (closest to professor's curiosity focus and located within the space of the corresponding curiosity zone) from the original dataset is taken and added to the already selected prototypes set of the professor; the same is done with the student's query. If it happens that both queries result to the same prototype, then the advantage to get it will be given to the student. If the largest curiosity zone does not have new points to be taken from the original dataset, then the second largest curiosity zone is examined, and so on. Stopping criteria: the algorithm stops when the curiosity of both actors will be completely satisfied, i.e.: either none of the curiosity zones (neither the student's nor the professor's ones) have vacant points in the original dataset; or the radius of the largest curiosity zone (simultaneously for the professor and for the student) reaches some predefined minimum $\varepsilon_0$. Otherwise, the stages (2), (3) and (4) are repeated continuously.

### *5.3. Generic settings for the experiments with the prototype selection algorithms*

As the key objective of prototype selection is to reduce the available dataset towards the most relevant instances for a potential and accurate classification tasks, there are two popular quality measures, which we are going to use for prototype selection methods evaluation: the Retention Rate (RR), which is calculated during the learning phase as the ratio between the amount of selected prototypes and the amount of instances in the original training set; and the Error Rate (ER), which is calculated during the testing phase as the ratio between the amount of misclassified instances and the total amount of instances.

We tested our algorithms on eight data sets of different complexity from the UCI machine-learning repository (Dua & Taniskidou, 2017). Details of individual datasets are presented in Table 1. We normalize the data (min-max scaling) and perform feature selection (dimensionality reduction) as a pre-processing step (towards adapting each dataset to be a two-dimensional one, since we restricted our previous considerations with the 2D ignorance zones). If the two best features (selected from the dataset with the univariate chi-squared $\chi^2$ statistical test) are representative, then we select them for the further experiments. Otherwise, the Principal Component Analysis (PCA) method is used to produce a pair of features well capturing the variance of the original dataset. Dimensionality reduction technique (PCA or $\chi^2$ test) and proportion of the variance explained by the selected principal components are presented in Table 1 for each dataset.

| Dataset | #Exemplars | #Attributes | #Classes | Type of dim. reduction | Proportion of variance |
|---|---|---|---|---|---|
| Iris | 150 | 4 | 3 | $\chi^2$ test | - |
| Wine | 178 | 13 | 3 | $\chi^2$ test | - |
| Pima | 768 | 8 | 2 | $\chi^2$ test | - |
| Breast Cancer | 699 | 9 | 2 | PCA | 0.76 |
| Ionosphere | 351 | 34 | 2 | PCA | 0.42 |
| Glass | 214 | 9 | 7 | PCA | 0.63 |
| Bupa | 345 | 6 | 2 | PCA | 0.6 |
| Transfusion | 748 | 4 | 2 | PCA | 0.93 |

Table 1. Description of the datasets adapted and used for the experiments.

During all the experiments, the 10-fold cross-validation is applied to each dataset by dividing it into ten equal parts and, in turn, using one block as a testing set and the remaining nine as the training set. Cross-validation is repeated ten times and results for each dataset are calculated as the averages over one hundred experiments.

The proposed IPS and APS algorithms and the experiments, validating their efficiency and effectiveness, were implemented in Python 2.7. We used NumPy package (numpy.org) for efficient management of multi-dimensional arrays and basic functionality of linear algebra. Classification and dimensionality reduction were implemented with the help of open source SciPy (scipy.org) and Scikit-learn (scikit-learn.org) packages. Matplotlib (matplotlib.org) and Seaborn (seaborn.pydata.org) libraries were used for visualization.

We use both the arithmetic and the contraharmonic means for averaging the algorithms' performance estimations during the experiments. We added the contraharmonic mean because (opposite to the traditional arithmetic average) it aims to mitigate the impact of small outliers and aggravate the impact of the large ones. The contraharmonic mean of positive real numbers $(x_1, x_2, \ldots, x_n)$ is as follows:

$$C(x_1, x_2, \ldots, x_n) = \frac{x_1^2 + x_2^2 + \cdots + x_n^2}{x_1 + x_2 + \cdots + x_n}.$$

The contraharmonic mean is always greater than the arithmetic mean and both means are equal only if the averaging numbers are equal to each other. This makes it a fair, strict and trustful metrics to aggregate evaluations for the qualitatively negative characteristics of the tested algorithms (such as error or retention rates) because it provides more punishment for the worst cases of the algorithm performance than traditional arithmetic average. Therefore, in some experiments, in addition to the arithmetic mean, we also used the contraharmonic mean to average the algorithms' performance indicators for each dataset during the cross-validation process.

## 5.4. Results of the experiments

The first set of experiments aims to test the performance of the IPS algorithm. The goal was to check how well the selected prototypes represent the original dataset during the classification. Therefore, we made three groups of tests: when all the dataset has been used as a prototype set during testing; when the prototypes were the outcome of the IPS algorithm with a rectangular domain boundary; and when the prototypes were the outcome of the IPS algorithm with a circular domain boundary. Table 2 contains the results, which are estimations for the error and retention rates, which correspond to each group of experiments with each of eight datasets. One can see that the IPS algorithm (for both rectangular and circular domains) makes quite compact selections for the prototype sets (comparably to the original sizes of the datasets) without essential loss of accuracy and, in many cases (also in average for all the datasets), even improves the classification accuracy. Notice also that the circular domain boundary for the datasets works better with the IPS algorithm than the rectangular domain does for the majority of the datasets.

| Dataset | Full set | | IPS (rectangular) | | IPS (circular) | |
|---|---|---|---|---|---|---|
| | ER | RR | ER | RR | ER | RR |
| Iris | 3.7 | 100 | 4.4 | 19.5 | 2.6 | 17.6 |
| Wine | 12.9 | 100 | 14.1 | 22.9 | 13.9 | 23.7 |
| Pima | 35.6 | 100 | 44.8 | 9.2 | 28.3 | 10.7 |
| Breast Cancer | 5.3 | 100 | 4.7 | 4.6 | 4.2 | 5.6 |
| Ionosphere | 29.0 | 100 | 28.4 | 18.1 | 19.9 | 18.7 |
| Glass | 38.4 | 100 | 48.7 | 40.2 | 42.7 | 45.8 |
| Bupa | 46.8 | 100 | 48.2 | 25.8 | 50.8 | 23.0 |
| Transfusion | 31.3 | 100 | 27.6 | 6.2 | 26.3 | 7.4 |
| AVERAGE | 25.38 | 100 | 27.61 | 18.31 | 24.84 | 19.06 |

Table 2. Results of the experiments with the Incremental Prototype Selection.

One of the experiments from Table 2 is presented visually in Figure 13, which shows the output of the IPS algorithm (circular domain) for the Wine dataset.

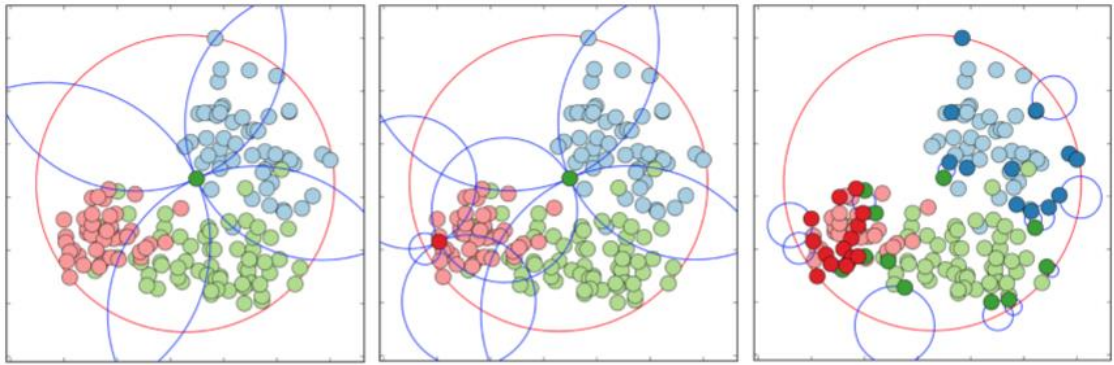

Figure 13. Incremental prototype selection algorithm with circular domain (Wine dataset) after the first, the second and the last iterations. Colored balls are data samples from the dataset and those bolder ones are from the set of already selected prototypes. Domain boundary is shown as a red circle and the blue circles are the voids, from which the ignorance zones are computed.

Let us check a potential benefit of our ignorance-aware incremental IPS algorithm comparably to some other popular methods. For the comparisons, we used modern realization of the two popular prototype selection algorithms: the Condensed Nearest Neighbor (CNN) *incremental* algorithm (Hart, 1968) and the Edited Nearest Neighbor (ENN) *decremental* algorithm (Wilson, 1972). CNN algorithm begins by randomly selecting one instance belonging to each output class from training set and putting them into the selection subset. Then each instance in the training set is classified using only the instances in the subset and it will be chosen for the selection, if misclassified. The process is repeated until there are no instances in the training set that are misclassified. ENN algorithm starts with the selection, which is equal to the whole training set, and then each instance in the selection is being removed if its class label does not agree with the majority of its *k* nearest neighbors.

Table 3 contains the results of the IPS vs. CNN vs. ENN comparisons. One may see that IPS performs essentially better than the competitors with the two major criteria (error and retention rates), i.e., in average it reduces number of selected prototypes and makes the training datasets almost twice more compact than CNN does and about three times more compact than ENN does. And, surprisingly, such (essentially reduced by IPS) datasets provide better accuracy (when tested with the 1-NN classifier) than the original complete datasets as well as the ones selected by CNN and ENN. For example, error rate with the datasets selected by IPS appeared to be 6.8 % better (smaller) than the one with the CNN selections, and 4.6 % better than the ENN provides. Interesting is that selections made by CNN and ENN provide worth accuracy than the complete datasets in our experiments, however, IPS selections happen to perform with 1-NN in a more accurate way (1.7%) than even the original complete datasets.

| Dataset | Full Set | | | IPS | | | CNN | | | ENN | | |
|---|---|---|---|---|---|---|---|---|---|---|---|---|
| | ER | RR | Time | ER | RR | Time | ER | RR | Time | ER | RR | Time |
| Iris | 11.0 | 100 | 6 | 11.2 | 18.2 | 277 | 12.2 | 36.4 | 35 | 9.2 | 94.5 | 8 |
| Wine | 17.6 | 100 | 7 | 17.1 | 28.0 | 1029 | 26.6 | 37.1 | 26 | 15.1 | 84.3 | 8 |
| Pima | 36.4 | 100 | 39 | 29.4 | 12.6 | 14550 | 36.8 | 59.0 | 409 | 38.5 | 69.0 | 42 |
| Breast Cancer | 6.3 | 100 | 41 | 5.1 | 5.7 | 1583 | 6.2 | 38.4 | 95 | 4.6 | 96.9 | 45 |
| Ionosphere | 31.3 | 100 | 18 | 27.3 | 25.9 | 10317 | 35.0 | 60.1 | 163 | 39.0 | 69.3 | 21 |
| Glass | 41.6 | 100 | 10 | 44.1 | 48.8 | 21202 | 59.3 | 17.3 | 71 | 46.5 | 33.7 | 13 |
| Bupa | 48.0 | 100 | 16 | 48.0 | 34.6 | 22967 | 51.8 | 70.2 | 163 | 50.2 | 57.3 | 18 |
| Transfusion | 32.2 | 100 | 35 | 28.7 | 14.9 | 20078 | 37.1 | 47.5 | 369 | 44.6 | 65.8 | 38 |
| AVERAGE | 28.05 | 100 | 21.5 | 26.36 | 23.59 | 11500 | 33.13 | 45.8 | 166 | 30.96 | 71.35 | 24 |

Table 3. Results of the performance comparison of the suggested ignorance-aware incremental prototype selection (IPS) algorithm vs. Condensed Nearest Neighbor (CNN) algorithm vs. Edited Nearest Neighbor (ENN) algorithm.

These advantages of the IPS algorithm are achieved on the expense of more thorough analysis and, therefore, essentially more processing time needs (comparably to CNN and ENN) at the prototype selection phase. This can be seen from the "Time" columns in Table 3. We did not target processing time in this paper because we want to get maximal value from the ignorance awareness specifically for the retention and error rates (meaning: the more time we spend at the offline selection phase, the less time we will need and less mistakes we will get at the online testing phase). The IPS algorithm can be optimized in future to decrease selection time, however, still it can be recommended when the dataset preparation for $k$-NN supposed to be performed offline in advance so that all the benefits (storage compactness, classification time and accuracy) will be visible during the online testing phase.

We also checked, what would be the gap in performance of a simple 1-NN classifier vs. some other more sophisticated classifiers, if applied to the same datasets. We try a Support-Vector-Machine (SVM) classifier and a multilayer perceptron (MLP) as kind of feedforward artificial neural network. Experiments with the same eight complete datasets show that in average SVM gives 3.85 % better accuracy that 1-NN, and MLP gives 2.52 % better accuracy than 1-NN. However, after applying our prototype selection (IPS) together with 1-NN we manage (with the more compact training datasets) to essentially lover the gap to the 2.16 % and 0.83 % accordingly.

In our next set of experiments, we assess the quality of prototypes, selected with the APS (Professor vs. Student) algorithm. The APS algorithm takes a complete set of training data as input, produces two sets of prototypes (selected by the student and the professor separately), and then these prototypes are used by the 1-NN classifier for testing. We made two experiments with two different prototype sets: the one chosen by the professor alone; and the aggregated (student + professor)'s set. The results of classification tests, shown in Table 4 (error rates) and Table 5 (retention rates), give evidence that the APS algorithm can be used as a self-sufficient prototype selection technique. One may see that the APS (both with the "professor's" prototype set only or with the "professor's + student's" prototype set) performs better than CNN and ENN and even gives better accuracy comparably to the complete datasets. Even with the combined "professor's + student's" prototype set, the APS gives better retention rate than the CNN and ENN do. Interesting that the "professor's" prototype set alone is enough to reach almost the same accuracy as our IPS algorithm, however, with essentially less number of prototypes (APS has about 6 % better retention rate than the IPS has), which can be observed in the Table 5. We added the contraharmonic averages to the Table 4 and Table 5 for the comparison aiming to show the APS advantages even with the bias towards the worst-case scenario.

| Dataset | Full set | | Professor's set | | Professor's + Student's set | |
|---|---|---|---|---|---|---|
| | Arithmetic Mean | Contra-harmonic Mean | Arithmetic Mean | Contra-harmonic Mean | Arithmetic Mean | Contra-harmonic Mean |
| Iris | 3.7 | 11.0 | 3.9 | 9.9 | 2.9 | 11.0 |
| Wine | 12.9 | 17.6 | 12.9 | 19.7 | 13.0 | 17.5 |
| Pima | 35.6 | 36.4 | 29.8 | 31.0 | 31.2 | 32.2 |
| Breast Cancer | 5.3 | 6.3 | 4.0 | 5.6 | 4.1 | 5.3 |
| Ionosphere | 29.0 | 31.3 | 26.4 | 29.6 | 25.8 | 28.3 |
| Glass | 38.4 | 41.6 | 47.1 | 50.1 | 38.4 | 41.5 |
| Bupa | 46.8 | 48.0 | 46.1 | 47.5 | 46.9 | 48.1 |
| Transfusion | 31.3 | 32.2 | 27.5 | 28.4 | 27.2 | 28.7 |
| AVERAGE | 25.38 | 28.05 | 24.71 | 27.73 | 23.69 | 26.58 |
| Compared with | AVERAGE (Full Set) | | 25.38 | 28.05 | 25.38 | 28.05 |
| | AVERAGE (IPS) | | 23.46 | 26.36 | 23.46 | 26.36 |
| | AVERAGE (CNN) | | 30.51 | 33.13 | 30.51 | 33.13 |
| | AVERAGE (ENN) | | 28.46 | 30.96 | 28.46 | 30.96 |

Table 4. Results of the experiments with the Adversarial Prototype Selection (Error Rate).

| Dataset | Full set | | Professor's set | | Professor's + Student's set | |
|---|---|---|---|---|---|---|
| | Arithmetic Mean | Contra-harmonic Mean | Arithmetic Mean | Contra-harmonic Mean | Arithmetic Mean | Contra-harmonic Mean |
| Iris | 100 | 100 | 16.5 | 16.8 | 40.0 | 40.3 |
| Wine | 100 | 100 | 21.9 | 22.8 | 52.2 | 52.8 |
| Pima | 100 | 100 | 10.4 | 11.2 | 26.9 | 28.0 |
| Breast Cancer | 100 | 100 | 5.9 | 6.1 | 13.8 | 14.1 |
| Ionosphere | 100 | 100 | 18.0 | 19.3 | 43.6 | 44.7 |
| Glass | 100 | 100 | 28.5 | 29.9 | 70.4 | 71.6 |
| Bupa | 100 | 100 | 20.8 | 23.5 | 53.5 | 56.3 |
| Transfusion | 100 | 100 | 8.0 | 9.1 | 21.4 | 22.8 |
| AVERAGE | 100 | 100 | 16.25 | 17.34 | 40.23 | 41.33 |
| Compared with | AVERAGE (Full Set) | | 100 | 100 | 100 | 100 |
| | AVERAGE (IPS) | | 22.31 | 23.59 | 22.31 | 23.59 |
| | AVERAGE (CNN) | | 45.71 | 45.80 | 45.71 | 45.80 |
| | AVERAGE (ENN) | | 71.26 | 71.35 | 71.26 | 71.35 |

Table 5. Results of the experiments with the Adversarial Prototype Selection (Retention Rate).

The observations above suggest the use of the IPS algorithm in cases when only the accuracy of further classification matters, however, we may suggest to use the APS algorithm when a good compromise between the accuracy and the compactness of the prototype set is desirable.

See, for example, Figure 14, which shows the output of the APS algorithm for the Wine dataset. Squares represent prototypes selected by the professor; triangles represent the choice of the student. Professor's sets contain exemplars mainly concentrated near the clusters' borders. It ignores the redundant unnecessary points and focuses on prototypes located next to the class contours. Based on the algorithm's logic, the examiner must prioritize the domain exploration by looking for prototypes in the largest ignorance zones. Selected exemplars comply with the professor's strategy: with knowledge about class contours, the professor is capable of asking hard questions and assessing student's knowledge accurately. Student's subset contains fewer exemplars in comparison with the professor's choice. The majority of the prototypes selected by the student are located inside the class contours defined by the professor. This disposition of points is a consequence of examinee's strategy to answer professor's questions correctly. Student's prototypes taken separately do not represent an independent value, but in conjunction with the professor's subset, they supplement the collective set of prototypes with valuable knowledge.

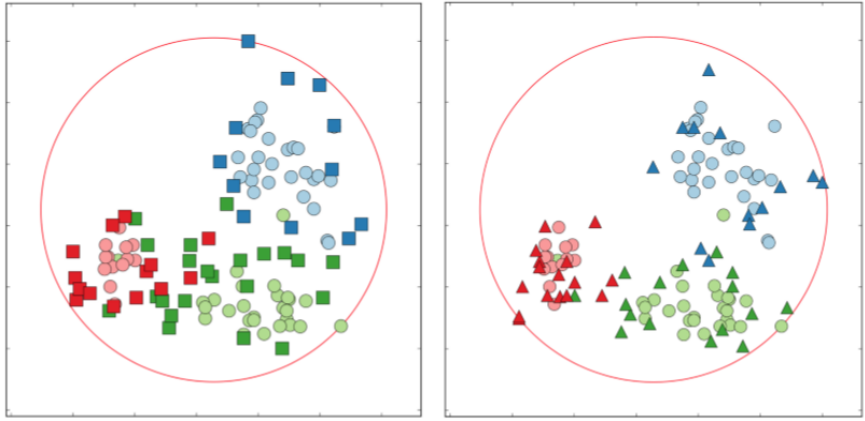

Figure 14. Adversarial prototype selection algorithm ("Student vs. Professor") with circular domain (Wine dataset) after the last iteration. Colored balls are data samples from the original dataset. Domain boundary is shown as a red circle. Bold squares are the prototypes selected by the professor and bold triangles are the prototypes selected by the student.

We checked how well our ignorance aware algorithms perform if they select prototypes not only for 1-NN but also for SVM and MLP classifiers. Additional experiments with Iris and Wine datasets confirm that (in the worth case scenario) SVM classifier is able to classify without loss of accuracy with less than 30% of the prototypes from the original datasets selected by our IPS or APS (particularly "Professor") algorithms. MLP can make the same (after IPS or APS) with 35% of the prototypes, while 1-NN requires only 25%, which makes it the main beneficiary of our algorithms, as we expected.

One can see that both algorithms (incremental and adversarial) are set up so that they can be generalized as such from a planar to a multidimensional space. However, in our experiments, to avoid computationally expensive hyperspheres' discovery, we suggest an approximation of IPS and APS for the higher dimensional cases, which we named as "*Quasi-Orthogonal Projections*" (QOP) algorithm. For a *n*-dimensional dataset with *k* data exemplars in it, the QOP algorithm guarantees the computational complexity to be not worth than $O(k^2 n)$.

Assume we have *n*-D (*n* > 2) dataset *S* with *n* attributes. QOP works as follows:
(1) Apply the Principal Component Analysis (PCA) and get the 2-D projection of *S*, which will be named as *S(0)*;
(2) $\forall i$: Remove attribute $i$ from *S*, obtain the reduced dataset, then use the PCA and get the 2-D projection of it named as *S(i)*;
(3) After applying step (2) $\forall i$, one gets *n* different "quasi-orthogonal" 2-D projections of *S*, and, therefore, the whole set of 2-D projections of *S* together with *S(0)* would be: {*S(0), S(1), S(2),…, S(n)*};
(4) $\forall S(i), i = \overline{0, n}$: apply our prototype selection algorithm (either IPS or APS) as described for 2-D analysis separately to each projection and get *n* + 1 sets of selected prototypes: {*P(0), P(1), P(2),…, P(n)*};

(5) Complete the final selection **P** of the prototypes from the original dataset *S* as follows: the sample $p_r$ from the dataset *S* will be included to **P**, if it appears at least *k* times within the prototype sets: {*P(0), P(1), P(2),..., P(n)*}, where the best *k* (1 ≤ *k* ≤ *n*) can be chosen experimentally (we recommend to use *k = n*, aiming the best Retention Rate).

(5*) Another option of the completion rule (5), which does not require any assumptions on the parameter *k*, could be as follows:
$$\mathbf{P} = P(0) \cap [P(1) \cup P(2) ... \cup P(n)].$$

We checked QOP (for both selection options: 5 and 5*) experimentally in 3D and noticed that adding a dimension improves the overall quality of selected prototype sets (by our IPS and APS algorithms) from about 1.5% to 12% for different datasets.

| IRIS-2D (with PCA) | Full Set | IPS | APS (S.) | APS (P.) | APS (P.+S.) |
|---|---|---|---|---|---|
| ER (Arithmetic) | 6.70 | 6.30 | 7.50 | 7.10 | **6.10** |
| ER (Contra-Harmonic) | 12.20 | 13.00 | 14.20 | 12.80 | **12.10** |
| RR (Arithmetic) | 100.00 | 23.90 | **22.30** | 30.00 | 52.30 |
| RR (Contra-Harmonic) | 100.00 | 24.70 | **22.90** | 30.50 | 53.00 |
| Ov. Quality (Arithmetic) | 46.65 | 84.90 | **85.10** | 81.45 | 70.50 |
| Ov. Quality (Contra-H.) | 43.90 | 81.15 | **81.45** | 78.35 | 67.45 |
| **IRIS-3D (with PCA)** | Full Set | IPS | APS (S.) | APS (P.) | APS (P.+S.) |
| ER (Arithmetic) | **6.40** | 11.10 | 18.90 | 10.90 | 7.30 |
| ER (Contra-Harmonic) | **12.20** | 18.20 | 34.40 | 18.30 | 13.50 |
| RR (Arithmetic) | 100.00 | 16.50 | **5.10** | 18.20 | 44.00 |
| RR (Contra-Harmonic) | 100.00 | 16.90 | **5.90** | 18.70 | 44.40 |
| Ov. Quality (Arithmetic) | 46.80 | 86.20 | **88.00** | 85.45 | 74.35 |
| Ov. Quality (Contra-H.) | 43.90 | **82.45** | 79.85 | 81.50 | 71.05 |
| **IRIS-3D* (with PCA)** | Full Set | IPS | APS (S.) | APS (P.) | APS (P.+S.) |
| ER (Arithmetic) | **6.40** | 8.10 | 8.50 | 6.90 | 6.80 |
| ER (Contra-Harmonic) | **12.20** | 15.30 | 16.50 | 13.80 | 12.50 |
| RR (Arithmetic) | 100.00 | 21.20 | **14.10** | 25.50 | 49.70 |
| RR (Contra-Harmonic) | 100.00 | 21.80 | **14.80** | 25.90 | 50.20 |
| Ov. Quality (Arithmetic) | 46.80 | 85.35 | **88.70** | 83.80 | 71.75 |
| Ov. Quality (Contra-H.) | 43.90 | 81.45 | **84.35** | 80.15 | 68.65 |

Table 6. Results of the experiments with Quasi-Orthogonal Projections algorithm in 3D applied to the Iris dataset. Here one can see the performance of IPS and APS algorithms in 2D and also (for comparison) there are results in 3D provided by two options of the QOP algorithm.

Table 6 shows the most challenging case of Iris dataset (where actually the classes are known to be well separated already in 2D). However, still our IPS and APS

("Professor + Student", "Student" and "Professor") managed to improve the overall quality in 3D due to tangible lowering the RR. We computed the ER and RR values for each algorithm and the overall quality for the algorithms as follows:

$$Overall\ Quality\ (\%) = 100 - \frac{\alpha \cdot ER + \beta \cdot RR}{\alpha + \beta},$$

where $\alpha$ and $\beta$ are the weights of importance of the ER and RR values respectively. In our experiments, we assume that $\alpha = \beta$, i.e., equal importance of the ER and RR for the overall quality estimation. From Table 6, one may see that the overall quality of prototype sets selected by all our ignorance aware algorithms improved in average as follows: 83.5% for 3D vs. 80.5% for 2D (with option (5) from the QOP); 78.7 % for 3D vs. 77.1% for 2D (with option (5) for the worst-case scenario); 82.4% for 3D vs. 80.5% for 2D (with option (5*)); 78.7 % for 3D vs. 77.1% for 2D (with option (5*) for the worst-case scenario). The QOP (being only a heuristic approximation of the actual *n*-D ignorance-aware prototype selection algorithms) gives a reasonable trade-off (precision vs. size) by essential lowering the RR with small raise on the ER.

*5.5. Experiments with the datasets in the GIS context*

In order to evaluate the performance of our IPS and APS algorithms also in the GIS context, we included the Forest Type Mapping Dataset (Johnson, Tateishi & Xie, 2012) to our experiments as the GIS-related dataset from the UCI machine-learning repository (Dua & Taniskidou, 2017). This dataset uses satellite imagery data of a forested area in Japan and the attributes contain the spectral values of satellite images together with the geographically weighted variables. The goal is to map different forest types using such geographically weighted spectral data (Johnson, Tateishi & Xie, 2012).

In our experiments with the Forest Type Mapping Dataset, we compare the performance of classification based on data reduced by our IPS and APS prototype selection algorithm and known ENN and CNN algorithms, as well as based on the original complete dataset without reduction.

Table 7 contains the results of the experiments. One can see that the prototype set collected by our IPS algorithm resulted to the most accurate classification (ER = 12.6) in average scenario, while our APS ("Professor + Student") provides the best average accuracy (ER = 16.7) for the worst-case scenario. The best average retention rates (RR = 22.7 / 23.3) has our APS ("Professor") algorithm, which also has the best overall quality (81.90 / 79.75) comparably to other algorithms.

| "Forest Types" Dataset | Full Set | CNN | ENN | IPS | APS ("Prof.") | APS ("Prof. + St.") |
|---|---|---|---|---|---|---|
| ER (Arithmetic) | 14.30 | 29.00 | 13.80 | **12.60** | 13.50 | 12.80 |
| ER (Contra-Harmonic) | 18.00 | 34.00 | 17.10 | 16.90 | 17.20 | **16.70** |
| RR (Arithmetic) | 100.00 | 26.90 | 76.30 | 34.10 | **22.70** | 56.00 |
| RR (Contra-Harmonic) | 100.00 | 27.00 | 76.40 | 34.40 | **23.30** | 56.60 |
| Overall Quality (Arithmetic) | 42.85 | 72.05 | 54.95 | 76.65 | **81.90** | 65.60 |
| Overall Quality (Contra-Harmonic) | 41.00 | 69.50 | 53.25 | 74.35 | **79.75** | 63.35 |

Table 7. Results of the experiments with the Forest Type Mapping Dataset.

For the further experiments, we decided to inject the GIS context into a couple of previously considered datasets from the UCI repository ("Iris" and "Wine"). We used the built-in approach (Eldawy & Mokbel, 2015), in which an existing non-spatial dataset is extended by injecting spatial awareness (Klippel, Hirtle & Davies, 2010) into it. In the original Iris dataset, there are three classes of irises with 50 samples in each class. Figure 15 illustrates the three intersected artificial spatial areas, within which all the 150 irises were randomly distributed so that their spatial coordinates has been captured and added to the dataset. As a result, we got the new dataset (named IRIS-GIS) for our further experiments. The same way we did to made spatial update of the Wine dataset to the WINE-GIS dataset. Our intention was to enable more challenging prototype selection cases for our IPS and APS algorithms vs. CNN and ENN algorithms within the GIS context.

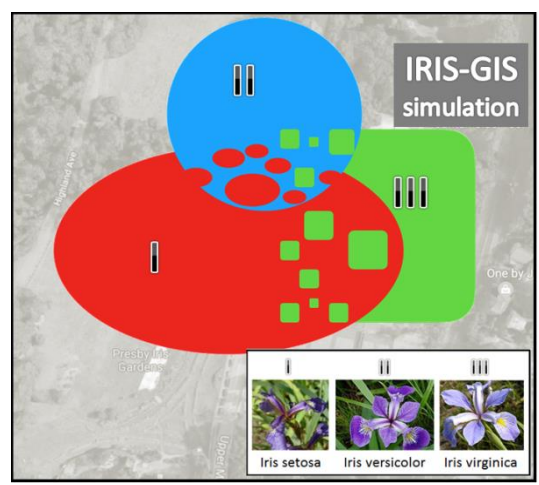

Figure 15. Injecting a spatial awareness into the Iris dataset. Samples of the three types of irises from the dataset were randomly placed within the three corresponding spatial areas (I, II and III) on the map. Obtained spatial coordinates has been added to the Iris dataset, which has been updated to the IRIS-GIS dataset as a result.

Table 8 and Table 9 show the results of intended comparisons (datasets with vs. without spatial awareness) for the "Iris vs. IRIS-GIS" and the "Wine vs. WINE-GIS" cases respectively. The datasets were preprocessed by the PCA for the dimensionality reduction. One can see that the overall quality of our IPS and APS ("Student") algorithms remains to be better than the quality of CNN and ENN for all the datasets. It is also interesting to notice that the injection of some challenging spatial awareness to the datasets made the APS ("Student") to be the winner among our algorithms (as well as among the others) both for the average and for the worst-case scenarios.

| IRIS (with PCA) | Full Set | CNN | ENN | IPS | APS (S.) | APS (Pr.) | APS (Pr.+S.) |
|---|---|---|---|---|---|---|---|
| ER (Arithmetic) | 6.70 | 8.80 | **5.50** | 6.30 | 7.50 | 7.10 | 6.70 |
| ER (Contra-Harmonic) | 12.20 | 14.90 | **11.60** | 13.00 | 14.20 | 12.80 | 12.10 |
| RR (Arithmetic) | 100.00 | 38.00 | 92.80 | 23.90 | **22.30** | 30.00 | 52.30 |
| RR (Contra-Harmonic) | 100.00 | 38.20 | 92.80 | 24.70 | **22.90** | 30.50 | 53.00 |
| Ov. Quality (Arithmetic) | 46.65 | 76.60 | 50.85 | 84.90 | **85.10** | 81.45 | 70.50 |
| Ov. Quality (Contra-H.) | 43.90 | 73.45 | 47.80 | 81.15 | **81.45** | 78.35 | 67.45 |
| IRIS-GIS (with PCA) | Full Set | CNN | ENN | IPS | APS (S.) | APS (Pr.) | APS (Pr.+S.) |
| ER (Arithmetic) | 4.20 | 5.30 | **3.50** | 6.10 | 4.40 | 4.60 | 4.00 |
| ER (Contra-Harmonic) | 9.40 | 10.80 | **9.00** | 14.00 | 9.10 | 11.70 | 9.30 |
| RR (Arithmetic) | 100.00 | 36.70 | 95.20 | **20.30** | 21.40 | 25.10 | 46.60 |
| RR (Contra-Harmonic) | 100.00 | 36.80 | 95.20 | **20.50** | 21.70 | 25.60 | 47.10 |
| Ov. Quality (Arithmetic) | 47.90 | 79.00 | 50.65 | 86.80 | **87.10** | 85.15 | 74.70 |
| Ov. Quality (Contra-H.) | 45.30 | 76.20 | 47.90 | 82.75 | **84.60** | 81.35 | 71.80 |

Table 8. Results of the experiments with the Iris vs. IRIS-GIS datasets.

| WINE (with PCA) | Full Set | CNN | ENN | IPS | APS (S.) | APS (Pr.) | APS (Pr.+S.) |
|---|---|---|---|---|---|---|---|
| ER (Arithmetic) | 3.60 | 7.50 | **2.90** | **2.90** | 4.40 | 3.80 | 3.50 |
| ER (Contra-Harmonic) | 8.40 | 15.10 | **7.00** | 7.20 | 9.40 | 8.10 | 8.80 |
| RR (Arithmetic) | 100.00 | 30.80 | 94.80 | 19.80 | **19.10** | 26.90 | 46.10 |
| RR (Contra-Harmonic) | 100.00 | 30.90 | 94.90 | 20.10 | **19.40** | 27.50 | 46.50 |
| Ov. Quality (Arithmetic) | 48.20 | 80.85 | 51.15 | **88.65** | 88.25 | 84.65 | 75.20 |
| Ov. Quality (Contra-H.) | 45.80 | 77.00 | 49.05 | **86.35** | 85.60 | 82.20 | 72.35 |
| WINE-GIS (with PCA) | Full Set | CNN | ENN | IPS | APS (S.) | APS (Pr.) | APS (Pr.+S.) |
| ER (Arithmetic) | 2.40 | 4.40 | **1.50** | 2.30 | 1.90 | 2.10 | 2.40 |
| ER (Contra-Harmonic) | 7.50 | 12.00 | **6.10** | 6.70 | 6.60 | 7.50 | 7.50 |
| RR (Arithmetic) | 100.00 | 29.40 | 96.20 | 20.60 | **18.80** | 27.00 | 45.80 |
| RR (Contra-Harmonic) | 100.00 | 29.50 | 96.30 | 20.70 | **19.00** | 27.60 | 46.20 |
| Ov. Quality (Arithmetic) | 48.80 | 83.10 | 51.15 | 88.55 | **89.65** | 85.45 | 75.90 |
| Ov. Quality (Contra-H.) | 46.25 | 79.25 | 48.80 | 86.30 | **87.20** | 82.45 | 73.15 |

Table 9. Results of the experiments with the Wine vs. WINE-GIS datasets.

Therefore, due to the new experiments, we once again observed the evidence that our ignorance-aware algorithms can successfully compete with other prototype selection algorithms applied on the datasets both with and without spatial awareness.

## 6. Conclusions

In this paper, we study the concept of ignorance, uncertainty and curiosity in the context of supervised learning. As the traditional knowledge discovery aims to capture hidden and potentially useful patters from data, we assume that the "ignorance discovery" might figure out potentially useful patterns from the voids within the data (ignorance areas or gaps). The latter patterns have a structure (boundary), which may indicate the potential curiosity areas useful for many applications. We have shown that many approaches are possible to discover the ignorance patterns, but we put the major focus on those, which are driven by the conflicts of tags (labels) of the dataset instances as well as by the structure of the domain boundary.

Our experimental study demonstrates the usefulness of the ignorance discovery for prototype selection from the datasets aiming compact and noise-resistant subset of data for future classification tasks. We suggested two algorithms (incremental and adversarial prototype selection) based on a curiosity-driven approach, in which the discovered ignorance zones indicate the curiosity focus for selecting every new prototype from the dataset. The logic of the algorithms is about finding such a subset of the dataset instances, which minimizes the size of ignorance area within the domain of selection. The adversarial version of the algorithm is a good illustration of the fact that the prototype selection is a process of the compromise search between two conflicting objectives: the compactness of knowledge used for future decisions vs. the capability to make further correct decisions with it. The experiments with real datasets show that the ignorance-awareness has a potential to improve the performance of the prototype selection methods. Possible variations of the algorithms depend on the choice of a suitable metrics (distance measure), some of which (Euclidean, min-max manifold distance, balanced view and social distance) are discussed in this paper.

In our experiments, we reduced the dimensionality of the datasets to 2D data domains and spaces, aiming to prove the concept of "beneficial ignorance" for potential GIS applications and beyond, yet indicating that the algorithms can be generalized aiming more generic multidimensional cases. However, we also suggested one possible way to deal with higher dimensions by using the heuristic QOP algorithm, which approximates the generic ignorance discovery case with reasonable efficiency and quality.

We considered just one potential use case for the ignorance discovery, which gives a clear benefit. However, our assumption that knowledge about the shape of ignorance can be at least as beneficial as known facts from the dataset still needs further research and experimental approval on a variety of domains and applications.

**References:**

Aurenhammer, F. (1991). Voronoi Diagrams – A Survey of a Fundamental Geometric Data Structure. *ACM Computing Surveys, 23*(3), 345–405. ACM New York, NY. doi:10.1145/116873.116880

Brighton, H., & Mellish, C. (2002). Advances in Instance Selection for Instance Based Learning Algorithms. *Data Mining and Knowledge Discovery*, *6*(2), 153–172. Kluwer Academic Publishers. doi:10.1023/A:1014043630878

Brunino, R., Trujillo, I., Pearce, F. R., & Thomas, P. A. (2007). The Orientation of Galaxy Dark Matter Haloes around Cosmic Voids. *Monthly Notices of the Royal Astronomical Society*, *375*(1), 184–190. doi:10.1111/j.1365-2966.2006.11282.x

Buchanan, M. (2018). Ignorance as Strength. *Nature Physics*, *14*, 428. doi:10.1038/s41567-018-0133-9

Chan, K. C., Hamaus, N., & Desjacques, V. (2014). Large-Scale Clustering of Cosmic Voids. *Physical Review D*, *90*(10), 103521. doi:10.1103/PhysRevD.90.103521

Chen, F., & Lu, C. T. (2008). Nearest Neighbor Query, Definition. In: *Encyclopedia of GIS* (pp. 776-782). Springer. doi:10.1007/978-0-387-35973-1_866

Couclelis, H. (2003). The Certainty of Uncertainty: GIS and the Limits of Geographic Knowledge. *Transactions in GIS*, *7*(2), 165–175. doi:10.1111/1467-9671.00138

De Bruin, S. (2008). Modelling Positional Uncertainty of Line Features by Accounting for Stochastic Deviations from Straight Line Segments. *Transactions in GIS*, *12*(2), 165-177. doi:10.1111/j.1467-9671.2008.01093.x

DeNicola, D. R. (2017). *Understanding Ignorance: The Surprising Impact of What We Don't Know*. MIT Press, Cambridge, MA.

Dua, D., & Taniskidou, E. K. (2017). *UCI Machine Learning Repository* [http://archive.ics.uci.edu/ml]. Irvine, CA: University of California, School of Information and Computer Science.

Eldawy, A., & Mokbel, M. F. (2015). The Era of Big Spatial Data: A Survey. *Information and Media Technologies*, *10*(2), 305-316. doi:10.11185/imt.10.305

Elzinga, D. J., & Hearn, D. W. (1972). The Minimum Covering Sphere Problem. *Management Science*, *19*(1), 96-104. doi:10.1287/mnsc.19.1.96

Ford, M. (2015). *Rise of the Robots: Technology and the Threat of a Jobless Future*. Basic Books.

Gabriel, K. R., & Sokal, R. R. (1969). A New Statistical Approach to Geographic Variation Analysis. *Systematic Biology*, *18*(3), 259–278. doi:10.2307/2412323

Garcia, S., Derrac, J., Cano, J., & Herrera, F. (2012). Prototype Selection for Nearest Neighbor Classification: Taxonomy and Empirical Study. *IEEE Transactions on Pattern Analysis and Machine Intelligence*, *34*(3), 417-435. doi:10.1109/TPAMI.2011.142

Goodfellow, I., Pouget-Abadie, J., Mirza, M., Xu, B., Warde-Farley, D., Ozair, S., Courville, A., & Bengio, Y. (2014). Generative Adversarial Nets. In: Z. Ghahramani *et al*. (Eds.), *Advances in Neural Information Processing Systems* (Vol. 2, pp. 2672–2680). MIT Press, Cambridge, MA.

Gupta, S., & Gupta, A. (2018). Handling Class Overlapping to Detect Noisy Instances in Classification. *The Knowledge Engineering Review*, *33*. Cambridge University Press. doi:10.1017/S0269888918000115

Hart, P. E. (1968). The Condensed Nearest Neighbour Rule. *IEEE Transactions on Information Theory*, 14(3), 515-516. doi:10.1109/TIT.1968.1054155

Johnson, B., Tateishi, R., & Xie, Z. (2012). Using Geographically-Weighted Variables for Image Classification. *Remote Sensing Letters*, *3* (6), 491-499. doi:10.1080/01431161.2011.629637

Kinkeldey, C. (2014). Development of a Prototype for Uncertainty-Aware Geovisual Analytics of Land Cover Change. *International Journal of Geographical Information Science*, *28*(10), 2076-2089. doi:10.1080/13658816.2014.891037

Klippel, A., Hirtle, S., & Davies, C. (2010). You-Are-Here Maps: Creating Spatial Awareness through Map-Like Representations. *Spatial Cognition & Computation*, *10*(2-3), 83-93. doi:10.1080/13875861003770625

Kononenko, I., & Kukar, M. (2007). *Machine Learning and Data Mining: Introduction to Principles and Algorithms*. Horwood Publishing Limited, Chichester, UK.

LeCun, Y., Bengio, Y., & Hinton, G. (2015). Deep Learning. *Nature*, *521*(7553), 436-444. doi:10.1038/nature14539

Leyk, S., Boesch, R., & Weibel, R. (2005). A Conceptual Framework for Uncertainty Investigation in Map-Based Land Cover Change Modelling. *Transactions in GIS*, *9*(3), 291-322. doi:10.1111/j.1467-9671.2005.00220.x

Mason, J., Klippel, A., Bleisch, S., Slingsby, A., & Deitrick, S. (2016). Special Issue Introduction: Approaching Spatial Uncertainty Visualization to Support Reasoning and Decision Making. *Spatial Cognition & Computation*, *16*(2), 97-105. doi:10.1080/13875868.2016.1138117

Minker, J. (1982). On Indefinite Databases and the Closed World Assumption. In: D. W. Loveland (Ed.), *6th Conference on Automated Deduction* (*Lecture Notes in Computer Science*, vol. 138, pp. 292–308). Springer, Berlin, Heidelberg. doi:10.1007/BFb0000066

Ogata, H., Hou, B., Li, M., Uosaki, N., & Mouri, K. (2013). Role of Passive Capturing in a Ubiquitous Learning Environment. In: *Proceedings of the IADIS International Conference Mobile Learning 2013, ML 2013* (pp. 117-124). IADIS.

Olvera-López, J. A., Carrasco-Ochoa, J. A., & Martínez-Trinidad, J. F. (2018). Accurate and Fast Prototype Selection based on the Notion of Relevant and Border Prototypes. *Journal of Intelligent & Fuzzy Systems*, *34*(5), 2923-2934. IOS Press. doi:10.3233/JIFS-169478

O'Sullivan, D., & Unwin, D. (2014). *Geographic Information Analysis*. John Wiley & Sons.

Pynchon, T. (1984). *Slow Learner. Early Stories* (Introduction, pp. 15-16). Boston: Little, Brown.

Reiter, R. (1981). On Closed World Data Bases. In: B. L. Lynn & N. J. Nilsson (Eds.), *Readings in Artificial Intelligence* (pp. 119–140). Elsevier.

Reuter, H. I., Nelson, A., & Jarvis, A. (2007). An Evaluation of Void- Filling Interpolation Methods for SRTM Data. *International Journal of Geographical Information Science*, *21*(9), 983-1008. doi:10.1080/13658810601169899

Ritter, J. (1990). An Efficient Bounding Sphere. In: A. S. Glassner (Ed.), *Graphics Gems* (pp. 301-303). Academic Press Professional, San Diego, CA.

Silver, D., Hubert, T., Schrittwieser, J., Antonoglou, I., Lai, M., Guez, A., Lanctot, M., Sifre, L., Kumaran, D., Graepel, T., & Lillicrap, T. (2017). Mastering Chess and Shogi by Self-Play with a General Reinforcement Learning Algorithm. *arXiv preprint arXiv:1712.01815*

Terziyan, V. (2017). Social Distance Metric: From Coordinates to Neighborhoods. *International Journal of Geographical Information Science*, *31*(12), 2401-2426. doi:10.1080/13658816.2017.1367796

Turing, A. M. (1950). Computing Machinery and Intelligence. *Mind*, *59*(236), 433-460. doi:10.1093/mind/LIX.236.433

Warwick, K., & Shah, H. (2017). Taking the Fifth Amendment in Turing's Imitation Game. *Journal of Experimental & Theoretical Artificial Intelligence*, *29*(2), 287-297. doi:10.1080/0952813X.2015.1132273

Wilson, D. L. (1972). Asymptotic Properties of Nearest Neighbor Rules Using Edited Data. *IEEE Transactions on Systems, Man and Cybernetics*, *2*(3), 408-421. doi:10.1109/TSMC.1972.4309137

Yuan, M. B., Butternfield, M., Gahegan, M., & Miller, H. (2005). Geospatial Data Mining and Knowledge Discovery. In: R. B. McMaster & E. L. Usery (Eds.), *Research Challenges in Geographic Information Science* (Chapter 14, pp. 365–388). John Wiley & Sons, CRC Press, Boca Raton, FL.

Zubek, J., & Kuncheva, L. (2018). Learning from Exemplars and Prototypes in Machine Learning and Psychology. *arXiv preprint arXiv:1806.01130*